\documentclass[11pt]{mystyle}

\usepackage{xcolor}
\usepackage{mathtools,bm}
\usepackage{algorithm}
\usepackage{algpseudocode}
\usepackage{array,multirow,booktabs,tabularx}
\usepackage{float}
\usepackage{placeins}
\usepackage{xspace}
\usepackage[numbers,sort&compress]{natbib}
\usepackage[nameinlink,capitalise,noabbrev]{cleveref}
\usepackage{authblk}

\newcommand{\cM}{\mathcal{M}}
\newcommand{\cS}{\mathcal{S}}
\newcommand{\cA}{\mathcal{A}}
\newcommand{\cO}{\mathcal{O}}

\newcommand{\cH}{\mathcal{H}}

\DeclareMathOperator*{\argmin}{arg\,min}
\newcommand{\codemodel}{Code as Model\xspace}
\newcommand{\arcworld}{\textsc{Code-as-Model}\xspace}
\newcommand{\den}[1]{[\![#1]\!]}               %
\newcommand{\comp}{\mathcal{C}}                %
\newcommand\blfootnote[1]{\begingroup\renewcommand\thefootnote{}\footnotetext{#1}\endgroup}

\theoremstyle{definition}
\newtheorem{definition}{Definition}

\theoremstyle{plain}

\newcommand{\ourrhae}{100.0}
\newcommand{\baseonerhae}{99.0}      %
\newcommand{\nooarhae}{85.1}         %
\newcommand{\opinerhae}{78.4}        %
\newcommand{\dtrhae}{38.1}           %
\newcommand{\chrhae}{20.5}           %

\newcommand{\arcstandardgap}{59.3}   %
\newcommand{\numgames}{25}
\newcommand{\numceiling}{25}         %
\newcommand{\ouractions}{7{,}518}    %
\newcommand{\humanactions}{17{,}135} %
\newcommand{\actionratio}{0.44}      %
\newcommand{\actionpercentage}{44\%} %
\newcommand{\baseactions}{8{,}347}   %
\newcommand{\nooaactions}{11{,}697}  %
\newcommand{\opineactions}{12{,}859} %
\newcommand{\dtactions}{9{,}789}     %
\newcommand{\chactions}{14{,}558}    %

\definecolor{oursbg}{RGB}{236,242,249}
\definecolor{barfill}{RGB}{47,102,166}
\definecolor{bartrack}{RGB}{224,227,232}
\newlength{\barw}
\newcommand{\ratiobar}[1]{\makebox[\barw][l]{%
  \textcolor{bartrack}{\rule[-0.22ex]{\barw}{1.05ex}}\hspace{-\barw}%
  \textcolor{barfill}{\rule[-0.22ex]{#1\barw}{1.05ex}}}}

\runningtitle{Memento 3: Model-based RSI
}
\correspondingauthor{Jun Wang}
\title{\textsc{Memento~3}: Model-Based Recursive \\ Self-Improvement through Reflective Rulebooks}

\DeclareRobustCommand{\paperauthorblock}{%
  {\normalfont\normalsize\bfseries
  Haoyu Zhao$^{1}$ \quad Zhengxu Yu$^{2}$ \quad Zhiyuan He$^{2}$ \quad Meng Fang$^{3}$\\[1pt]
  Rasul Tutunov$^{2}$\quad
  Haitham Bou-Ammar$^{1,2}$ \quad Weilin Luo$^{2}$ \quad Jun Wang$^{1,\dagger}$\par}
  \vspace{4pt}
  {\normalfont\fontsize{10}{12}\selectfont
  \makebox[\linewidth][c]{$^{1}$University College London\quad $^{2}$Huawei Noah's Ark Lab, UK\quad $^{3}$University of Liverpool}\par}
  \vspace{2pt}
}
\author{\paperauthorblock}
\date{}

\begin{document}

\begin{abstract}
Learning to act in unfamiliar environments requires agents to infer how the world works and revise that understanding as new evidence arrives. Yet limited observations can support multiple world models that explain past interactions but predict different outcomes in unseen states. We introduce \textsc{Memento~3}, building on the Memento series to enable frozen LLM agents to continually learn explicit world models through external memory. The agent maintains a natural-language \emph{rulebook} as persistent semantic memory, recording revisable hypotheses about environment dynamics while leaving unknown aspects underspecified. It compiles this rulebook into executable code for prediction and planning. Through a continual loop of \emph{observation}, \emph{reflection}, \emph{rule revision}, \emph{compilation}, and \emph{verification}, the agent uses prediction errors to refine both the rulebook and its code. Updated code is accepted only when the LLM judges it faithful to the rulebook and cell-exact replay reproduces the observed transitions.
We investigate this process as a model-based route to recursive self-improvement (RSI): the agent autonomously explores the environment, revises its world model, and uses verified updates to guide subsequent interaction and learning, while the underlying LLM remains fixed. A population extension maintains multiple world models in parallel, sharing interaction evidence and using their predictions to guide exploration. On ARC-AGI-3, the single-model agent clears every level of all $\numgames$ public games, achieves a mean Relative Human Action Efficiency (RHAE) of $\ourrhae$, and uses \actionpercentage{} of the human action count. In an Atari Pong case study, a learned feedback controller wins $21\!:\!0$ in each of three evaluated episodes with different openings, without further LLM calls.
\end{abstract}

\maketitle
\blfootnote{$^{\dagger}$Corresponding author: \texttt{jun.wang@cs.ucl.ac.uk}}
\thispagestyle{plain}
\pagestyle{plain}

\begingroup
\setlength{\intextsep}{2pt}
\begin{figure}[H]
  \centering
  \captionsetup{skip=4pt}
  \includegraphics[width=\linewidth]{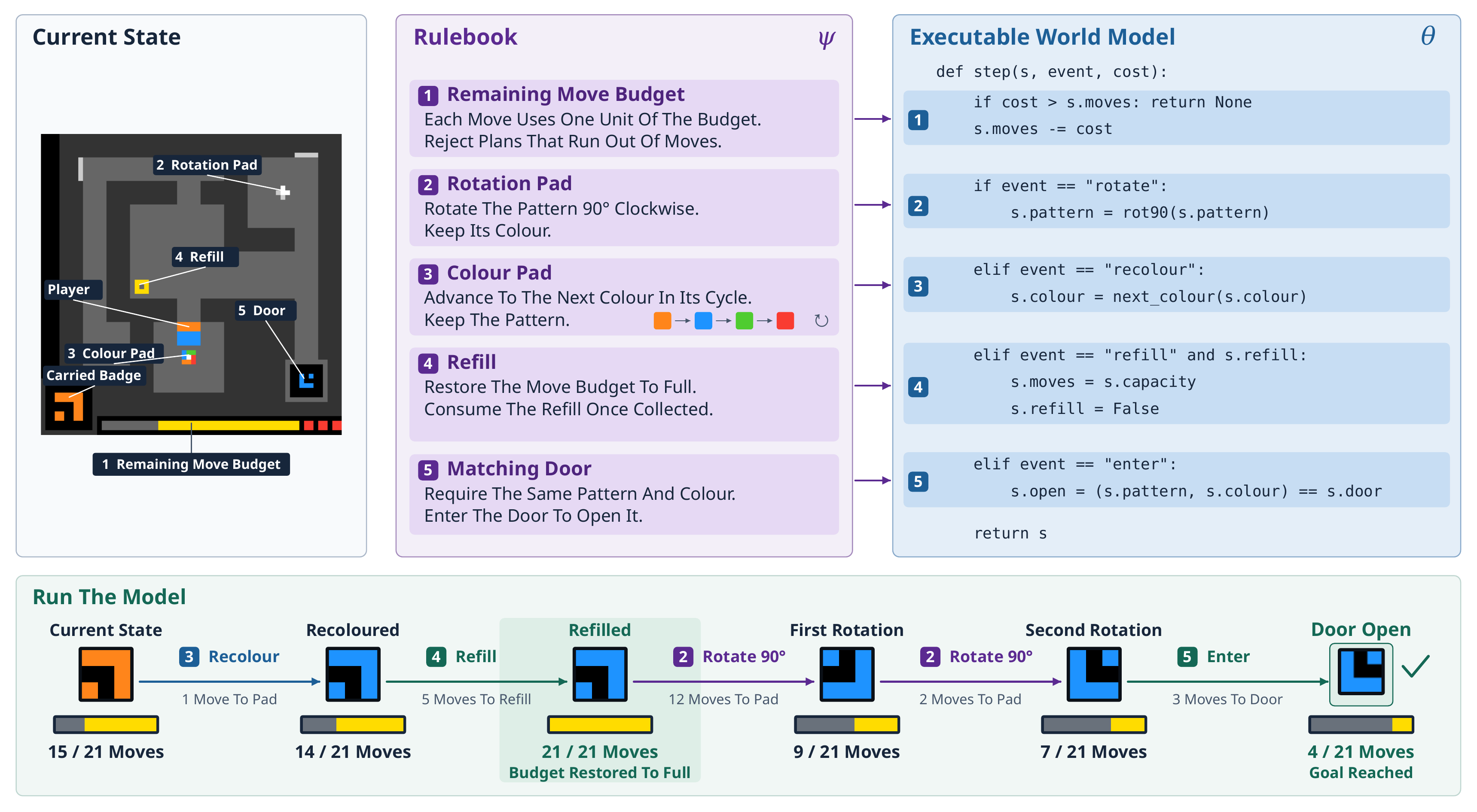}
  \caption{\textbf{From a rulebook to an executable world model.} Numbered rules map to
  corresponding branches of a simplified transition function. A planner uses the model to predict
  a route that changes the badge's pattern and colour, collects a refill, and reaches the matching
  door within the move budget.}
  \label{fig:rulebook-mini}
\end{figure}
\endgroup

\clearpage

\section{Introduction}
\label{sec:intro}

Recursive self-improvement aims to enable autonomous agents to acquire knowledge
from experience and progressively improve their capabilities.
An agent entering an unfamiliar environment must learn not only which actions are useful, but also
what its actions do. It may initially know neither the environment's objects and dynamics nor the
criterion for success. A model-based agent learns a world model to predict the consequences of
actions before executing them~\citep{sutton1991dyna,ha2018worldmodels,chua2018,schrittwieser2020muzero}.
Programs are attractive representations for such models, encoding state transitions and goal
conditions as executable functions that a planner can query repeatedly. They support fast,
inspectable simulation, while discrepancies between predicted and observed transitions provide
concrete counterexamples to the current model. Recent code world models demonstrate the promise
of learning such programs from interaction~\citep{tang2024worldcoder,dainese2024gifmcts,rodionov2026executable,opineworld}.

Agreement with observed transitions, however, does not guarantee correct generalisation. A finite
interaction history typically leaves a version space of programs that reproduce every observed
transition but disagree on unobserved states~\citep{mitchell1977version}. When planning queries such
states, the selected program must make predictions that the available evidence does not determine.
For example, if every observed wall lies at column 7, the hypotheses ``a box stops at a wall'' and
``a box stops at column 7'' may explain the same trajectory while predicting different outcomes when
a wall appears elsewhere. Replay can reject an inconsistent program, but cannot identify the correct
rule among consistent alternatives. The agent therefore needs to maintain and revise its
generalisation hypotheses as new evidence arrives.

Our Memento series~\citep{zhou2025memento,wang2025memento2,zhou2026mementoskills} provides a complementary perspective on this problem by treating external memory, rather than model parameters, as the agent's evolving learning state. Memento introduced continual adaptation through episodic memory, in which interaction outcomes are written into memory and retrieved to improve later decisions~\citep{zhou2025memento}. Memento~2 formalised this process as stateful reflective learning, interpreting memory writing as policy evaluation and memory reading as policy improvement~\citep{wang2025memento2}. Memento-Skills then extended the learning state from individual experiences to reusable procedural memory, allowing agents to construct and refine executable skills without updating the underlying LLM~\citep{zhou2026mementoskills}. These works primarily improve the policy side of an agent by learning which decisions or procedures should be used. An unfamiliar interactive environment introduces the complementary problem of learning the model side: the rules that determine how the world behaves.

We introduce \textsc{Memento~3} as the model-based continuation of the Memento series.
\textsc{Memento~3} extends reflective learning from policy and procedural memory to explicit world models.
Its central design, which we call \codemodel, represents the agent's evolving understanding of the
environment through a natural-language rulebook and an executable realisation.
The interaction history provides episodic evidence, while a natural-language \emph{rulebook}
serves as semantic memory, recording the agent's current hypotheses about objects, actions,
dynamics, and goals.
The rulebook makes these hypotheses explicit so that later observations can be used to assess and
revise them. Code is generated from the rulebook as its executable realisation for prediction and
planning, as illustrated in \Cref{fig:rulebook-mini}. The two representations serve complementary roles:
the rulebook states the environment rules the agent believes, while execution fixes how those rules
are applied. Together they form a persistent world-model hypothesis that can be inspected, tested,
and revised throughout interaction.

This design lets us investigate a model-based route to \emph{recursive self-improvement (RSI)}.
RSI describes an AI system autonomously designing and developing its own successor, with subsequent
versions continuing this process~\citep{favaro2026aibuildsitself}. Its recursive mechanism is a feedback loop
in which the system uses its current capabilities to improve the components supporting its
reasoning, planning, and learning~\citep{weng2026harness}.
In \codemodel, the persistent rulebook and its executable realisation are both objects of revision
and tools for further learning. They guide subsequent planning and exploration, shaping the evidence
the agent obtains for its next round of model revision. Their rules and predictions also provide a
basis for diagnosing prediction errors.
Through the reflective learning loop in \Cref{fig:overview}, the agent adopts verified updates to
these components while the underlying LLM remains fixed.

Exact replay can leave several world-model hypotheses consistent with the same evidence. The LLM
favours concise rulebooks and implementations, subject to rulebook fidelity and exact replay, as a
practical minimum-description-length bias~\citep{rissanen1978modeling}. This preference does not
eliminate uncertainty, and committing to one hypothesis may steer exploration away from evidence
that would falsify it. We therefore extend the single-model setting to $N$ world models maintained
in parallel, each represented by a rulebook--executable pair and sharing the same interaction
history. One member is sampled for each planning episode, and the resulting transitions are used
to check all members. Different models can thus guide exploration while remaining consistent
with shared evidence; $N=1$ recovers the single-model setting.

We evaluate \codemodel on ARC-AGI-3~\citep{arcagi3}, where agents must infer unfamiliar game rules
and goals under a human-derived action budget. The single-model agent clears every level of all
$\numgames$ public games, reaches the Relative Human Action Efficiency ceiling with a mean RHAE of
$\ourrhae$, and uses \actionpercentage{} of the corresponding human action count.
Compared with Claude Opus~5 with the ARC Prize Standard
harness~\citep{arcprize2026opus5}, our agent achieves a $\arcstandardgap$-point higher mean RHAE.
We also examine whether the learning mechanism can support feedback control
in addition to discrete planning. In Atari Pong, the learned model supports a controller that wins
$21\!:\!0$ in three evaluated episodes with different openings, acting without further LLM calls.

\paragraph{Contributions.}
\begin{itemize}\itemsep2pt
\item We introduce \textsc{Memento~3}, extending the Memento series to semantic world-model memory
through \codemodel. A continual loop of observation, reflection, rule revision, compilation, and
verification updates the rulebook and its executable realisation. The verified model guides
subsequent interaction while the underlying LLM remains fixed.
\item We extend this learning loop to multiple world models maintained in parallel. The models
share interaction evidence for updating and verification, allowing different models to guide
exploration.
\item We evaluate \textsc{Memento~3} on ARC-AGI-3, including targeted comparisons of the rulebook
and population extension. The single-model agent clears all $\numgames$ public games using
\actionpercentage{} of the human action count. An Atari Pong case study further demonstrates a
learned feedback controller that wins $21\!:\!0$ in three evaluated episodes without LLM calls
during execution.
\end{itemize}

\section{Model-Based Recursive Self-Improvement}
\label{sec:formulation}

\begin{figure}[!t]
  \centering
  \includegraphics[width=\linewidth]{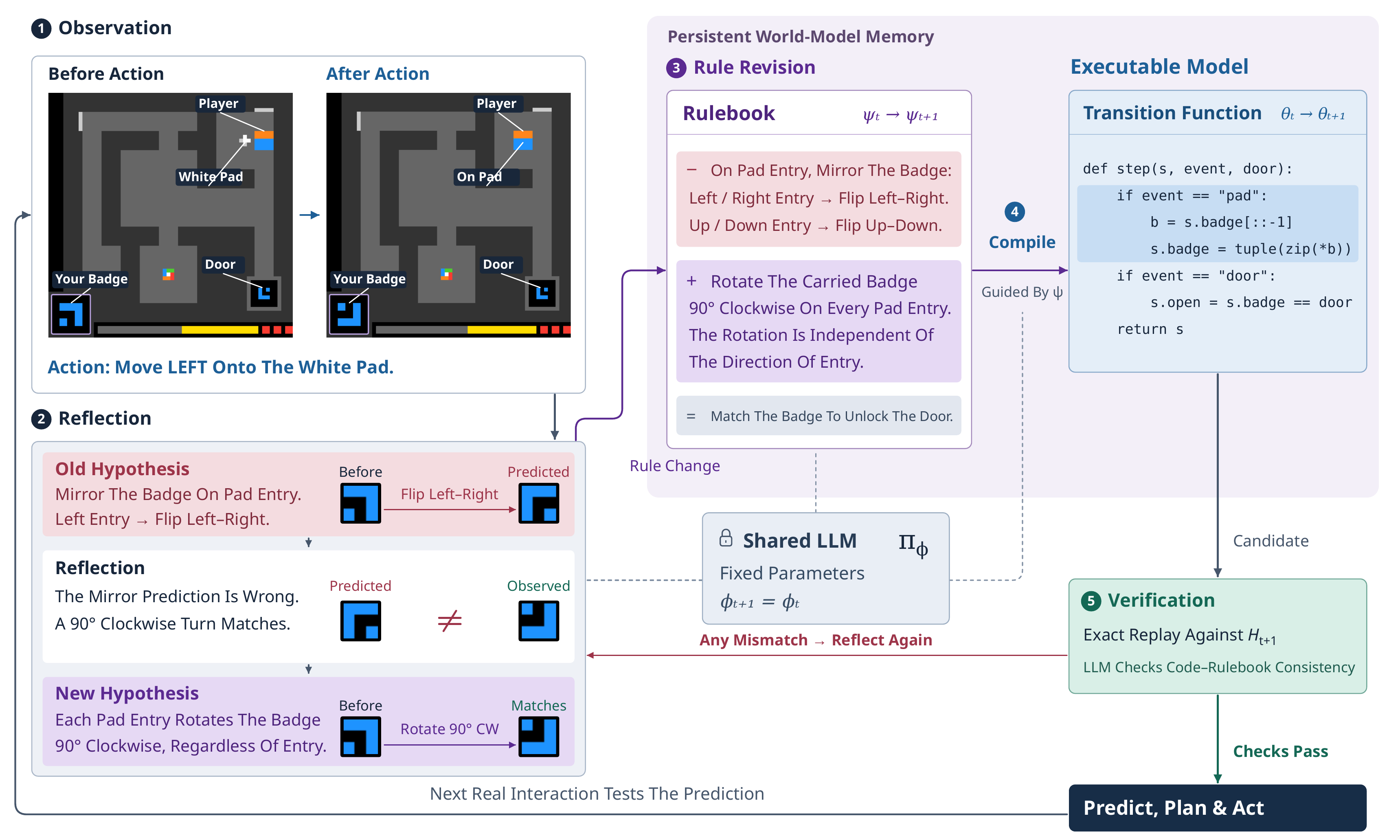}
  \caption{\textbf{Continual world-model learning with a frozen LLM.} The agent follows a continual
  loop of \emph{observation}, \emph{reflection}, \emph{rule revision}, \emph{compilation}, and
  \emph{verification}. In this example, an observed badge change contradicts a mirror hypothesis.
  Reflection yields a clockwise-rotation rule, which is recorded in the rulebook and compiled into
  code. Verification checks the executable against the interaction history and the rulebook.
  The accepted model then guides further interaction, producing evidence for the next revision.
  Here $H_t$, $\psi_t$, and $\theta_t$ denote the interaction record, rulebook, and executable after
  $t$ real actions; $\pi_\phi$ denotes the LLM's output distribution with fixed parameters $\phi$.}
  \label{fig:overview}
\end{figure}

 Learning a world model from interaction involves forming and revising hypotheses about an unfamiliar environment's rules as new evidence arrives. In \codemodel{}, the agent records its current hypotheses in a natural-language
rulebook and implements them in executable code for prediction and planning.
Figure~\ref{fig:overview} summarises this continual learning loop: new observations drive revisions
to the rulebook and executable, and the revised model guides subsequent interaction.

\subsection{Problem   formulation }
\label{sec:problem}

  \paragraph{Environment and interaction record.}
In our setting, an agent aims to reach a goal within a limited action budget.
It infers the environment's transition rules and goal conditions through interaction.
We model the  agent's interaction with   this unknown environment $\cM$  as a deterministic, goal-directed partially observable
Markov decision process (POMDP)~\citep{kaelbling1998planning},
\begin{equation}
  \begin{aligned}
  \cM&=\langle\cS,\cA,\cO,T,\rho,G\rangle,\\
  T&:\cS\times\cA\to\cS,
  \qquad \rho:\cS\to\cO,
  \qquad G\subseteq\cS.
  \end{aligned}
  \label{eq:environment}
\end{equation}
Here $\cS$ is the underlying state space, $\cA$ the action space, and $\cO$ the
observation space;  $G$ is the unknown set of goal states. The unknown transition function $T$
acts on the complete environment state, whereas $\rho$ returns the observation available to the agent.

At interaction step $t$, the environment is in state $s_t\in\cS$. The agent receives
observation $o_t=\rho(s_t)\in\cO$ and selects an action $a_t\in\cA$. The agent  observes the action interface and interaction
feedback, but has no access to the environment source, transition rules, underlying environment
state, or a natural-language task description. The environment also reports goal completion,
providing evidence about the goal without revealing its full specification. Direct evidence about
the environment comes from these interactions; the agent may use prior knowledge to interpret it.

 Let $\cH$ denote the space of finite interaction records.  For a single interaction trajectory,
 $H_t\in\cH$ denotes  the action--observation history   after  $t$   real actions:
 \begin{equation}
  H_t=(o_0,a_0,o_1,\ldots,a_{t-1},o_t),
  \qquad o_i=\rho(s_i).
  \label{eq:history}
\end{equation}
 Here $H_i$ denotes the interaction history up to and including observation $o_i$.  Associated completion feedback is retained with
this record and left implicit in the notation.

\paragraph{Belief over states and environment rules.}
The environment tuple $\cM$ specifies how the world evolves, while a belief represents
the agent's uncertainty about its hidden state and unknown rules given the interaction history.
We formulate learning and control in this unknown environment as a
Bayes-adaptive POMDP~\citep{ross2007bayesadaptive}.
Let $\mathfrak M$ be a discrete family of candidate models
$m=(T_m,\rho_m,G_m)$ on common discrete state, action, and observation spaces.
The true model $m^\star=(T,\rho,G)\in\mathfrak M$ is fixed but unknown.
The augmented hidden state is therefore $(s_t,m^\star)$. Given a prior over the initial state
and model, its belief is
\begin{equation}
  b_t(s,m)=\Pr(s_t=s,m^\star=m\mid H_t).
  \label{eq:joint-belief}
\end{equation}
Here $b_t$ is a probability distribution over $\cS\times\mathfrak M$; its initial value
conditions the prior on the initial feedback. The environment state $s_t$ evolves
under the true transition function $T_{m^\star}=T$. The model $m^\star$ remains fixed
as the agent revises its hypotheses about the environment rules.
For the belief equations, write $y_t=(o_t,c_t)$, where
$c_t=\mathbf 1[s_t\in G_{m^\star}]$ is the reported completion signal retained in $H_t$.
The indicator $\mathbf 1[\cdot]$ equals one when its condition holds and zero otherwise.

\paragraph{World-model hypothesis and learning state.}
We adopt a model-based approach that learns an environment model from interaction
and uses its predictions to select actions. In \codemodel{}, a frozen LLM proposes transition rules
and goal conditions in a natural-language rulebook and compiles them into executable code.
A planner queries the verified executable to search for action sequences predicted to reach a goal
within the remaining budget. The agent follows the resulting plan, using discrepancies between
predicted and observed outcomes to guide revisions to the rulebook and code.

Let $\Psi$ be the space of finite natural-language rulebooks and $\Theta$ the space of executable
world-model programs. The rulebook $\psi_t\in\Psi$ serves as persistent semantic memory, recording the agent's
current hypotheses about entities, action semantics, dynamics, goals, and termination conditions. Its executable realisation
$\theta_t\in\Theta$  supplies a transition predictor and an inferred goal condition over its  own
 internal state space.   The persistent world-model hypothesis is the pair $h_t=(\psi_t,\theta_t)$.
Together with the interaction record, it forms the agent's external learning state
\begin{equation}
  Z_t=(H_t,\psi_t,\theta_t).
  \label{eq:external-state}
\end{equation}
The interaction record $H_t$ provides evidence, the rulebook $\psi_t$ expresses hypotheses that
generalise beyond that evidence, and the executable $\theta_t$ implements those hypotheses for
prediction, replay verification, and planning.

A useful rulebook states transferable environment rules
with enough precision to constrain their implementation in code.
For any rulebook $\psi\in\Psi$, its \emph{denotation} $\den{\psi}\subseteq\Theta$ is the set of programs
that faithfully implement its stated hypotheses. A rulebook can leave unobserved mechanics
unspecified, admitting multiple executable realisations. The rulebook carries rule-level reasoning and
revision. An implementation correction may change $\theta_t$ without changing $\psi_t$, whereas a change
in the agent's understanding of the environment must be recorded in $\psi_t$ before it is
compiled into code.

The agent learns by revising the rulebook and executable stored in $Z_t$, while the
underlying LLM remains fixed.
Let $\pi_\phi$ denote the conditional output distribution of an LLM with parameter vector $\phi$.
At each interaction step $t$, its parameters $\phi_t$ satisfy
\begin{equation}
  \phi_{t+1}=\phi_t=\phi.
  \label{eq:frozen-llm}
\end{equation}

The LLM generates candidate rules and code from the interaction record and the current
hypothesis. Its pretrained knowledge guides this search. A revision can change the
proposed entities, internal state variables, transition rules, or goal conditions.

The evolving world model serves both as the object of revision and as a tool for further learning.

\begin{definition}[Model-based recursive self-improvement]
\label{def:model-based-rsi}
Model-based recursive self-improvement is a process in which an agent autonomously
revises a persistent world-model hypothesis $h_t$ to improve subsequent prediction and
decision-making. The agent uses $h_t$ to predict the consequences of candidate actions
and guide planning and exploration. These predictions provide a basis for interpreting
feedback and diagnosing errors. The agent uses the resulting evidence to construct a
revised hypothesis $h_{t+1}$, which then guides subsequent interaction and further model revision.
\end{definition}

\paragraph{Prediction through execution.}
To predict the consequences of actions, the executable $\theta\in\Theta$ specifies
an internal state space $\widehat{\cS}_\theta$, an initialiser $\iota_\theta$, a transition
predictor $f_\theta$, an observation function $\rho_\theta$, and a goal set:
\begin{equation}
  \begin{aligned}
  \iota_\theta&:\cO\to\widehat{\cS}_\theta,
  &f_\theta&:\widehat{\cS}_\theta\times\cA\to\widehat{\cS}_\theta,\\
  \rho_\theta&:\widehat{\cS}_\theta\to\cO,
  &\widehat G_\theta&\subseteq\widehat{\cS}_\theta.
  \end{aligned}
  \label{eq:cwm}
\end{equation}
The set $\widehat G_\theta$ identifies internal states predicted to satisfy the
goal. The agent uses completion feedback to infer the goal condition; $\rho_\theta$ predicts the
observations used for replay verification.
For an accepted hypothesis $h=(\psi,\theta)$, these functions and $\widehat G_\theta$
realise the rules in $\psi$ as a candidate environment model with internal state space
$\widehat{\cS}_\theta$.

Let $\widehat s_i^\theta\in\widehat{\cS}_\theta$ denote the internal state after replaying
the first $i$ actions of a trajectory. Starting from its initial observation, replay gives
\begin{equation}
  \widehat s_0^\theta=\iota_\theta(o_0),
  \qquad \widehat s_{i+1}^\theta=f_\theta(\widehat s_i^\theta,a_i),
  \qquad \widehat s_\theta(H_i):=\widehat s_i^\theta.
  \label{eq:model-replay-state}
\end{equation}

The initialiser selects an internal state consistent with the initial observation.
Replay advances that state through the recorded actions, retaining information that may not
be visible in the latest observation. The reconstructed state is the candidate model's
estimate of the current environment state and supplies the starting point for prediction and planning.
 After replaying $H_i$ and applying action $a_i$,
the program's  next-observation prediction    is
\begin{equation}
  \widehat o_\theta(H_i,a_i)
  =\rho_\theta\!\left(f_\theta(\widehat s_\theta(H_i),a_i)\right).
  \label{eq:model-observation}
\end{equation}
    Syntactically different programs may therefore induce the same observable predictions.

Whenever the executable is revised, its internal
state is reconstructed by replay under the revised program.
For multiple trajectories, $H_t$ denotes their accumulated record. Each trajectory is replayed
from its own initial observation, and the consistency and verification conditions apply to
every recorded transition within each trajectory.
In this case, $\widehat s_\theta(H_t)$ denotes the reconstructed state of the current trajectory.

\paragraph{Belief approximation for planning.}
The joint belief $b_t$ defined in Eq.~\eqref{eq:joint-belief} describes uncertainty
over the environment state and model given the interaction history.
To approximate this belief for planning, the agent uses the frozen LLM to propose and revise
rulebook--executable hypotheses from the interaction record. Verification tests their
observation predictions against the evidence, and replay reconstructs their internal states.
In the single-model case, the agent retains one accepted hypothesis
$h_t=(\psi_t,\theta_t)$. Replaying the interaction history $H_t$ under $\theta_t$
reconstructs the current internal state $\widehat s_{\theta_t}(H_t)$.
The resulting point belief approximation is
\begin{equation}
  \widehat b_t
  =\delta_{(h_t,\widehat s_{\theta_t}(H_t))}.
  \label{eq:point-hypothesis}
\end{equation}
Here $\delta$ denotes the point-mass distribution concentrated on the selected hypothesis and its
reconstructed state in $\widehat{\cS}_{\theta_t}$.
The belief approximation $\widehat b_t$ is a probability distribution that
assigns probability one to this pair and zero to all other hypothesis--state pairs.
The single-model planner evaluates action sequences under the selected
hypothesis $h_t$, starting from the reconstructed state $\widehat s_{\theta_t}(H_t)$.
The external learning state $Z_t$ retains both $h_t$ and the full history $H_t$,
allowing the LLM to revisit the evidence when revising the hypothesis.
Replay under an accepted revision
reconstructs the current internal state and updates $\widehat b_t$.
Section~\ref{sec:map} extends this point belief approximation to a population of
model--state hypotheses.

\paragraph{Continual  rulebook and  executable world-model learning.}
Given black-box interaction access to $\cM$, a fixed LLM, and a budget of  $B$   real actions, construct
an online update  operator  $U$ for the rulebook and executable and a control operator $P$ for action
selection. The first pair $h_0$ is constructed from the initial record $H_0=(o_0)$.
At each step, $P$ uses the current learning state $Z_t$ to select an action, and $U$ uses
the resulting evidence to update the stored pair:
\begin{equation}
  \begin{aligned}
  a_t&\sim P(\cdot\mid Z_t),\\
  s_{t+1}&=T(s_t,a_t),\qquad o_{t+1}=\rho(s_{t+1}),\\
  h_{t+1}&\sim U(\cdot\mid H_{t+1},h_t).
  \end{aligned}
  \label{eq:update-control}
\end{equation}
Here $H_{t+1}$ extends $H_t$ with the new action--observation pair. The control operator plans
using $\theta_t$ and may select exploratory actions when no plan is available. The update operator
retains the current pair,  revises the   rulebook and compiles it into code, or repairs the existing
implementation. These operators may be randomised.   A pair is adopted for planning only if its executable faithfully implements the rulebook
and reproduces every recorded next observation when replaying the recorded actions:
\begin{equation}
  \theta_t\in\den{\psi_t},
  \qquad \widehat o_{\theta_t}(H_i,a_i)=o_{i+1}\quad\text{for all }i<t.
  \label{eq:problem-consistency}
\end{equation}
Section~\ref{sec:loop} describes how these requirements are checked through an LLM assessment
of rulebook fidelity and exact replay of recorded interactions.

\begin{samepage}
The objective is to reach a goal within the interaction budget. Let $\tau_G$ be the first
step at which the environment enters $G$, and define the success probability $p_B$ and
conditional interaction cost $c_B$ by
 \begin{equation}
  \begin{aligned}
    \tau_G&=\inf\{t\geq 0:s_t\in G\},
    \qquad p_B=\Pr(\tau_G\leq B),\\
    c_B&=\mathbb E[\tau_G\mid\tau_G\leq B]\quad (p_B>0),
  \end{aligned}
  \label{eq:problem-objective}
\end{equation}
\end{samepage}
with $\tau_G=+\infty$ if no goal is reached. The primary objective is to maximise the probability
$p_B$ of reaching a goal within budget; among procedures with equal positive success probability,
the secondary objective is to minimise the conditional action cost $c_B$~\citep{teichteil2012safest,kolobov2012deadends}. Probability and expectation
are over the agent's internal randomness for a
fixed environment and initial state. Every real action, including exploration and failed
attempts, counts towards the budget; model construction, simulated rollouts, and replay do not.
 The goal set and unit action costs specify this objective directly; no separate scalar reward
function is required.

\begin{figure}[t]
  \centering
    \includegraphics[width=0.96\linewidth]{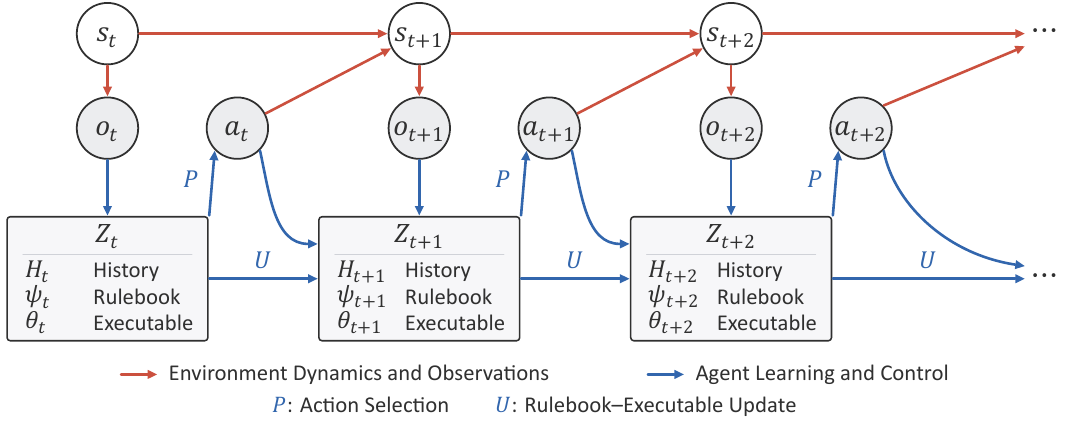}
   \caption{\textbf{A graphical model of recursive self-improvement in Code as Model.}
  The underlying environment state $s_t$ produces observation $o_t$, which enters the external learning
  state $Z_t=(H_t,\psi_t,\theta_t)$. The agent selects $a_t$ from $Z_t$, and the environment
  advances to $s_{t+1}$. The next observation $o_{t+1}$, together with $Z_t$ and $a_t$,
  determines the information available for updating $Z_{t+1}$.
  Rectangles contain the history, rulebook, and executable; white circles denote hidden
  states and grey circles observations and actions. $P$ selects actions, and $U$ updates the
  rulebook--executable pair. Ellipses indicate continued interaction.
  Completion feedback also enters the interaction history but is omitted from the diagram for clarity.}
  \label{fig:problem-setting}
\end{figure}

A finite   record  can admit several   rulebook--executable pairs  that agree on observed   interactions
 but differ elsewhere.   The current executable  guides action selection and therefore influences the evidence
  the agent obtains next. This evidence can in turn trigger revisions to the rulebook and its implementation.
 Continual executable world-model learning thus couples model
induction, exploration, and goal-directed planning.
Figure~\ref{fig:problem-setting} presents a graphical model of this coupling,
showing how observations and actions connect the evolving environment state with the agent's
external learning state. Red arrows represent
the environment dynamics $T$ and observation function $\rho$; blue arrows represent history
extension,   updates to the rulebook--executable pair  under $U$, and action selection under $P$.

\subsection{Reflective rule revision, compilation, and verification}
\label{sec:operator}
\label{sec:loop}

After action $a_t$, feedback $y_{t+1}$ provides evidence about the current state
and environment rules. The inference task is to update the joint belief in
Eq.~\eqref{eq:joint-belief} using this evidence.
For feedback $y=(o,c)$, define the deterministic likelihood under model $m$ as
\begin{equation}
  \ell_m(y\mid s')=
  \mathbf 1[o=\rho_m(s')]\,
  \mathbf 1[c=\mathbf 1[s'\in G_m]].
  \label{eq:feedback-likelihood}
\end{equation}
Let $p(y\mid b,a)$ be the predictive probability of feedback $y$. The ideal Bayesian
belief update is
\begin{equation}
  \begin{aligned}
  p(y\mid b,a)&=\sum_{m,s}b(s,m)\,
     \ell_m(y\mid T_m(s,a)),\\
  b_{t+1}(s',m)&=
  \frac{\ell_m(y_{t+1}\mid s')
    \sum_s\mathbf 1[s'=T_m(s,a_t)]\,b_t(s,m)}
  {p(y_{t+1}\mid b_t,a_t)}.
  \end{aligned}
  \label{eq:belief-update}
\end{equation}
The sums range over the discrete state and model spaces. The numerator first
propagates each possible state through its candidate dynamics and then retains
states and models compatible with the new feedback. The denominator normalises
the result. For positive predictive probability, write the update as
$b_{t+1}=\tau(b_t,a_t,y_{t+1})$.

\codemodel approximates this inference by constructing and selecting
rulebook--executable hypotheses from the interaction record.
The frozen LLM expresses candidate rules in a rulebook and compiles them into a
simulator with its own internal state representation. Replay tests the simulator's
observation predictions against the history and reconstructs its current internal
state. An accepted rulebook--executable pair and its reconstructed state supply
the belief approximation $\widehat b_t$ used for planning.

Each interaction advances the external state through   the five stages illustrated in
Figure~\ref{fig:overview}. Together, these stages implement
the update operator $U$ for the rulebook--executable pair $h_t$, using the frozen LLM
$\pi_\phi$ to diagnose errors and revise rules or code. The proposed pair is checked against
the two requirements in Eq.~\eqref{eq:problem-consistency} before it is used for planning.

\begin{enumerate}
  \item \textbf{Observation.} Executing $a_t$ produces observation $o_{t+1}$,
        which extends the available record as
        \begin{equation}
          H_{t+1}=H_t\mathbin{\Vert}(a_t,o_{t+1}).
          \label{eq:observe}
        \end{equation}
         Here $\Vert$ appends the action--observation pair to the current trajectory in the record.

  \item \textbf{Reflection.} Let $\mathcal D$ be the set of revision choices: retaining the current
        rulebook--executable pair $(\psi_t,\theta_t)$, revising the
         rulebook and its executable realisation, or repairing the executable while
        retaining the rulebook. Given the updated interaction history and the current pair, the frozen
        LLM proposes a revision choice $d_{t+1}\in\mathcal D$:
        \begin{equation}
          \begin{aligned}
            \mathcal D=\{\mathsf{retain},\mathsf{revise\_rulebook},\mathsf{repair\_code}\},\ \ d_{t+1}\sim\pi_\phi(\cdot\mid H_{t+1},\psi_t,\theta_t).
          \end{aligned}
          \label{eq:diagnose}
        \end{equation}

A prediction mismatch can arise from an incorrect environment rule or from code
        that fails to implement the intended rule.

  \item \textbf{Rule revision.} Choosing to revise the rulebook invokes the   frozen LLM's revision  operator
        $\Delta_\phi$; otherwise the   rulebook  is retained. The proposed rulebook $\widetilde\psi_{t+1}$ is

\begin{equation}
          \widetilde\psi_{t+1}=
          \begin{cases}
            \Delta_\phi(\psi_t;H_{t+1}), & d_{t+1}=\mathsf{revise\_rulebook},\\
            \psi_t, & \text{otherwise}.
          \end{cases}
          \label{eq:revise}
        \end{equation}

The proposed rulebook constrains the next executable through
        $\den{\widetilde\psi_{t+1}}$.

  \item \textbf{Compilation.}

When reflection requests rule revision or code repair, a rulebook-guided compiler
$\comp_\phi$ proposes an executable realisation $\widetilde\theta_{t+1}$ from the
previous executable, the proposed rulebook, and the updated interaction record.
We use $\bot$ when no previous executable exists:

\begin{equation}
          \comp_\phi:(\Theta\cup\{\bot\})\times\Psi\times\cH
          \rightsquigarrow\Theta,
          \qquad
          \widetilde\theta_{t+1}\sim
          \comp_\phi(\cdot\mid\theta_t,\widetilde\psi_{t+1},H_{t+1}),
          \label{eq:compile}
        \end{equation}

For $d_{t+1}=\mathsf{retain}$, set $\widetilde\theta_{t+1}=\theta_t$ and proceed
directly to verification.

Each compiled program $\theta$ supplies a simulator for its candidate model. From an
        internal state $\widehat s\in\widehat{\cS}_{\theta}$ and action $a$, it generates
        \begin{equation}
          \widehat s'=f_\theta(\widehat s,a),\qquad
          \widehat o'=\rho_\theta(\widehat s'),\qquad
          \widehat c'=\mathbf 1[\widehat s'\in\widehat G_\theta].
          \label{eq:compiled-simulator}
        \end{equation}
        Here $\widehat c'$ is predicted goal completion.

  \item \textbf{Verification.} Exact replay   checks the observation-consistency  requirement
        in Eq.~\eqref{eq:problem-consistency}. For a single trajectory,   let
        $\mathrm{verify}(\theta;H_t)$ indicate whether $\theta$ predicts every recorded next observation correctly,
        and let $\Theta_t$ denote the set of programs that pass this check:
         \begin{equation}
          \mathrm{verify}(\theta;H_t)
          =\prod_{i<t}\mathbf 1[\widehat o_\theta(H_i,a_i)=o_{i+1}],
          \qquad
          \Theta_t=\{\theta\in\Theta:\mathrm{verify}(\theta;H_t)=1\}.
          \label{eq:verify}
        \end{equation}
        Let $\mathcal V_t$ denote the admissible pairs whose executable both passes replay
        and faithfully implements its rulebook. The proposal  is accepted only if   it belongs
        to this set for the updated record:
        \begin{equation}
          \mathcal V_t=\{(\psi,\theta)\in\Psi\times\Theta:
            \theta\in\Theta_t\cap\den{\psi}\},
          \qquad
          (\widetilde\psi_{t+1},\widetilde\theta_{t+1})\in\mathcal V_{t+1}.
          \label{eq:editselect}
        \end{equation}
          On acceptance, the proposed pair becomes $h_{t+1}=(\psi_{t+1},\theta_{t+1})$,
        giving $Z_{t+1}=(H_{t+1},\psi_{t+1},\theta_{t+1})$ for the next
        round of prediction and action selection. Otherwise, the  agent returns to reflection,
        revision, and compilation until it produces an admissible pair.

Conditional on the recorded initial observations and actions and the program's
        single initialisation, replay defines an observation likelihood
        $L_t^{\mathrm{obs}}(h)=\mathrm{verify}(\theta;H_t)$ for $h=(\psi,\theta)$.
        It removes hypotheses assigning zero likelihood to a recorded observation.
        Each new proposal is checked against the entire record, so a repair must explain
        earlier evidence as well as the latest discrepancy.

\end{enumerate}

For   deterministic executables that terminate, finite replay    yields an exact pass-or-fail result:
the recorded action sequences are executed, and each predicted next observation is compared with
the corresponding recorded observation. Rulebook fidelity,
$\theta\in\den{\psi}$, is a semantic condition and is assessed by the frozen LLM from the rulebook
and source. Acceptance therefore combines an LLM assessment of semantic fidelity with an exact
execution check; it is not based on language-model confidence alone.   Selecting hypotheses based on language-model rankings can exclude correct candidates~\citep{wang2024hypothesis}, which motivates grounding
acceptance in execution wherever the consequences are observable.

In the single-model case, replay under the accepted executable reconstructs the
current internal state. Together with the selected rulebook--executable pair,
this state gives the updated point belief approximation
\begin{equation}
  h_{t+1}\sim U(\cdot\mid H_{t+1},h_t),\qquad
  \widehat b_{t+1}
  =\delta_{(h_{t+1},\widehat s_{\theta_{t+1}}(H_{t+1}))}.
  \label{eq:implemented-belief-update}
\end{equation}

\subsection{Model selection and planning}
\label{sec:single-selection}
\label{sec:plan}

Verification can leave multiple rulebook--executable pairs in the admissible set $\mathcal V_t$.
Selecting a working hypothesis completes the update operator $U$; the control operator
$P$ then uses that hypothesis to choose actions. In the single-model case, \codemodel
uses the LLM to prefer a simpler explanation and implementation among the admissible
pairs. We denote this LLM-guided revision and selection by
$\operatorname{Simplify}_{\mathrm{LLM}}$.

 Writing $\mathcal L(\psi)$ for rulebook description length and $\mathcal L(\theta\mid\psi)$ for
  the description length of the executable given the rulebook, the idealised selection objective
 for $t\geq1$  is
\begin{equation}
  (\psi_t,\theta_t)
  =\operatorname{Simplify}_{\mathrm{LLM}}(H_t,\psi_{t-1},\theta_{t-1})
  \approx
  \argmin_{(\psi,\theta)\in\mathcal V_t}
  \bigl[\mathcal L(\psi)+\mathcal L(\theta\mid\psi)\bigr].
  \label{eq:map}
\end{equation}

The objective favours the shortest combined description among admissible
rulebook--executable pairs. To make its maximum a posteriori (MAP) interpretation explicit,
assume a proper prior $\mu_0$ over rulebook-faithful pairs with
$\mu_0(\psi,\theta)\propto
2^{-\mathcal L(\psi)-\mathcal L(\theta\mid\psi)}$.
Under the observation likelihood defined in verification, the posterior over
rulebook--executable pairs is
\begin{equation}
  \mu_t(h)=
  \frac{\mu_0(h)L_t^{\mathrm{obs}}(h)}
       {\sum_{\bar h}\mu_0(\bar h)L_t^{\mathrm{obs}}(\bar h)},
  \qquad h=(\psi,\theta),
  \label{eq:hypothesis-posterior}
\end{equation}
provided its normalising constant is positive. A shortest admissible pair maximises
$\mu_t$. With the executable's fixed initialiser, each hypothesis $h=(\psi,\theta)$
also determines the replayed state $\widehat s_\theta(H_t)$. The corresponding belief
over hypotheses and their internal states is
$\sum_h\mu_t(h)\delta_{(h,\widehat s_\theta(H_t))}$.
Selecting one pair and its reconstructed state for planning yields the point
representation in Eq.~\eqref{eq:point-hypothesis}.

The frozen LLM approximates the selection in Eq.~\eqref{eq:map} by proposing simpler
environment rules and corresponding source. Each proposed simplification is checked
again for rulebook fidelity and replay consistency.

\paragraph{Belief-space control and executable planning.}
The Bayesian counterpart of the success objective in Eq.~\eqref{eq:problem-objective}
averages over the states and models in the joint belief of Eq.~\eqref{eq:joint-belief}.
Let $J_k(b)$ be the largest probability of reaching a goal within $k$ remaining
actions, for belief $b$.
For this finite-horizon calculation, take completed states to be absorbing. Then
\begin{equation}
  \begin{aligned}
  J_0(b)&=\sum_{s,m}b(s,m)\,\mathbf 1[s\in G_m],\\
  J_k(b)&=\max_{a\in\cA}\sum_y p(y\mid b,a)\,
        J_{k-1}\!\left(\tau(b,a,y)\right),\qquad k\geq1.
  \end{aligned}
  \label{eq:belief-control}
\end{equation}
The sum covers possible observation--completion pairs with positive predictive
probability. A belief-based controller selects a maximising action, with $k=B-t$.
Conditional action cost
breaks ties between policies with equal positive success probability.

Actions affect both the environment and the evidence available for subsequent decisions.
In Eq.~\eqref{eq:belief-control}, each possible feedback $y$ leads to an updated belief
$\tau(b,a,y)$, on which future action choices depend. When competing rule hypotheses
predict different outcomes for an action, the observed outcome can distinguish them
and improve subsequent planning. An information-seeking action can therefore increase
the probability of reaching a goal even when it makes little immediate progress.
The continuation horizon $k-1$ accounts for the action spent obtaining this evidence.

For control, the single-model agent uses the point belief $\widehat b_t$ defined in
Eq.~\eqref{eq:point-hypothesis}. Conditional on this selected model and state,
future transitions and observations are deterministic. Planning under this
certainty-equivalent approximation reduces to finding a goal-reaching action
sequence within the remaining budget.

An accepted executable induces the internal planning model
$\widehat{\cM}_\theta=\langle\widehat{\cS}_\theta,\cA,f_\theta,\widehat G_\theta\rangle$.
For this deterministic model with unit action costs, define $V_\theta(x)$ as the
length of a shortest finite path from $x\in\widehat{\cS}_\theta$ to $\widehat G_\theta$
under $f_\theta$, with $V_\theta(x)=+\infty$ if no such path exists. A trajectory that
never reaches the goal, including one trapped in a cycle under an improper policy, has
infinite cost. These model-conditioned shortest-path costs satisfy
      \begin{equation}
  V_\theta(x)=
  \begin{cases}
    0, & x\in\widehat G_\theta,\\
    1+\min_{a\in\cA}V_\theta(f_\theta(x,a)), & \text{otherwise}.
  \end{cases}
  \label{eq:bellman}
\end{equation}

  From a state with  $0<V_\theta(x)<+\infty$,   repeatedly selecting a minimising action yields  a
finite goal-reaching sequence $\mathrm{plan}_\theta(x)$. Each action decreases $V_\theta$ by one, so this
sequence cannot cycle. At a predicted goal the plan is empty; if $V_\theta(x)=+\infty$,
no goal-reaching plan is returned. Planning starts from $\widehat s_\theta(H_t)$, the internal state reconstructed
from the current trajectory. For a nonempty plan, the  history-dependent policy   $\pi_\theta$
selects its first action:
 \begin{equation}
  \pi_\theta(H_t)
  =\operatorname{first}\!\left(\mathrm{plan}_\theta(\widehat s_\theta(H_t))\right).
  \label{eq:history-policy}
\end{equation}
These costs and plans are conditional on the current executable model and its reconstructed state.
  The control operator $P$ uses $\pi_\theta$ while following the plan; when  no plan is available,
  it explores or requests model revision.

Under the selected deterministic hypothesis, a shortest path of at most $B-t$ actions
predicts goal completion within budget at minimum action cost.

For exploration when no goal-reaching plan is available, the LLM uses the interaction
history and unresolved rulebook hypotheses to choose an intermediate target, such as
testing a mechanic, revealing an unseen region, or interacting with an unexplained object.
An auxiliary planner searches the current executable for an action sequence reaching
that target. Execution compares predicted and observed outcomes after each action and
stops on a mismatch, returning the new evidence to reflection.

Both planned and exploratory actions produce new observations, extending the record
in Eq.~\eqref{eq:observe}. This evidence tests the selected model's predictions and
enters the next reflective update in Eq.~\eqref{eq:implemented-belief-update}.
  The agent updates its own world-model components $(\psi_t,\theta_t)$,
which then guide subsequent action selection and reflection.
 This feedback loop underlies our model-based approach to RSI.

\subsection{Population extension: from one to many}
\label{sec:map}

The single-model planning approximation in Eq.~\eqref{eq:point-hypothesis} commits to one explanation
of the rules and current state. Other pairs in $\mathcal V_t$ can fit the same history
while prescribing different plans. With the LLM as the proposal and revision
operator, repeated runs can select different initial hypotheses, whose plans influence
the evidence observed later. A plausible but incorrect model can remain self-confirming
if its actions never expose its errors.

        For a fixed executable $\theta$, let   $R(\theta)\subseteq\cH$  be the histories reached while following
its plan-induced policy $\pi_\theta$. For the fixed true environment and initial state,
let $o^+(H,a)$ denote the next observation after history $H$ and action $a$. Over histories
where $\pi_\theta$ is defined, set
\begin{equation}
  E(\theta)=\{H:\widehat o_\theta(H,\pi_\theta(H))
  \neq o^+(H,\pi_\theta(H))\}.
  \label{eq:model-errors}
\end{equation}
        These are histories where the model makes an observable prediction error on its own
selected action. During this policy's execution, a model can remain unfalsified whenever
\begin{equation}
  E(\theta)\cap R(\theta)=\varnothing,
  \label{eq:selfconf}
\end{equation}
even if its predictions are wrong after other histories or actions. Rewriting the same hypothesis in language does not by itself remove this
self-confirming behaviour.

Model-based reinforcement learning represents uncertainty in environment
dynamics through probabilistic models or ensembles~\citep{kurutach2018,chua2018}. Posterior sampling
and related methods retain a sampled model or value function across an episode to support
temporally coherent exploration~\citep{osband2013,osband2016,luvanroy2017}.

  Let $N\geq1$ be the number of world-model hypotheses retained by the agent. We replace
the single pair in $Z_t$ with a population of admissible pairs,
 \begin{equation}
  \Phi_t=\{(\psi_t^i,\theta_t^i)\}_{i=1}^{N}\subseteq\mathcal V_t.
  \label{eq:invariant}
\end{equation}
The  resulting learning state consists of the shared record $H_t$ and the population $\Phi_t$.
The  $N=1$ case recovers the formulation above. For $N>1$, the system maintains $N$ world models
in parallel, each comprising a rulebook and its executable realisation. These models may encode
different hypotheses about the same environment. They share $H_t$ rather than interacting with
separate environment copies.

Write $h_t^i=(\psi_t^i,\theta_t^i)$ for the $i$th retained hypothesis and
$Z_t^{(N)}=(H_t,\Phi_t)$ for the population learning state.
The population induces an empirical belief approximation
\begin{equation}
  \widehat b_t^{(N)}
  =\frac{1}{N}\sum_{i=1}^{N}
    \delta_{(h_t^i,\widehat s_{\theta_t^i}(H_t))}.
  \label{eq:population-belief}
\end{equation}
Each term pairs a rulebook--executable hypothesis with its reconstructed state
in that executable's own state space. The equal weights describe the empirical
distribution over retained members; $N=1$ recovers Eq.~\eqref{eq:point-hypothesis}.
For action selection, the rule below samples uniformly over distinct nonempty plans,
giving retained alternatives opportunities to generate new evidence.

  Every real interaction extends this common record. All members are checked against the same
complete set of recorded trajectories.   An executable is falsified when its prediction contradicts the new observation. Each  affected
pair then undergoes the reflective update of Section~\ref{sec:loop}.

The population should preserve behavioural alternatives rather than syntactic variants of one program.
  At the shared history $H_t$, each model reconstructs its own internal state. Models returning
nonempty plans belong to the same class if those plans have identical action sequences:
      \begin{equation}
  \theta\sim_{H_t}\theta'
  \quad\Longleftrightarrow\quad
  \mathrm{plan}_\theta(\widehat s_\theta(H_t))
  =\mathrm{plan}_{\theta'}(\widehat s_{\theta'}(H_t)).
  \label{eq:behav-equiv}
\end{equation}
  Let  $\widehat\Phi_t$   denote the set of behavioural equivalence classes represented by population
members  with nonempty plans, and write $[\theta]$ for the class containing $\theta$.
When  $\widehat\Phi_t\neq\varnothing$, the system samples one class  uniformly  for each planning
episode. A representative $\theta$ supplies the action sequence $\bar a_{t:t+K-1}$ of length $K\geq1$:
\begin{equation}
  [\theta]\sim\mathrm{Uniform}(\widehat\Phi_t),
  \qquad \bar a_{t:t+K-1}=\mathrm{plan}_\theta(\widehat s_\theta(H_t)).
  \label{eq:sample-policy}
\end{equation}
       If no member supplies a nonempty plan and the task is unfinished, the control operator
explores or revises the models before sampling resumes.
 The selected member remains active
    and its sequence is followed until success, falsification, or replanning. Selecting different surviving models can expose the agent
to different action sequences and hence different histories.
The population is therefore designed to broaden model-guided exploration and reduce the dependence of
an entire run on one early draw.

The population learning loop consists of three operations. \textsc{Propose} uses the frozen LLM $\pi_\phi$
to generate a rulebook and compile its executable; the resulting pair is retained only if it
belongs to $\mathcal V_t$. \textsc{Repair} applies reflection, rule revision when required, compilation,
and verification to every falsified member. \textsc{Plan/Act} uniformly samples a surviving behavioural class,
executes the plan supplied by a representative executable, and adds each observed transition to the
shared interaction record. In \codemodel, the rulebooks make each member's rule hypotheses explicit,
so differences in predictions and plans can be examined at the level of environment rules.

The observation-based posterior in Eq.~\eqref{eq:hypothesis-posterior} makes the
remaining ambiguity explicit. Marginalising $\mu_0$ and $\mu_t$ over rulebooks
gives the corresponding prior and posterior over executables, denoted $p(\theta)$
and $p(\theta\mid H_t)$. Conditional on the recorded initial observations and actions,
every replay-consistent executable has unit observation likelihood. Thus, for any
$\theta,\theta'\in\Theta_t$ with positive prior probability,

\begin{equation}
  \frac{p(\theta\mid H_t)}{p(\theta'\mid H_t)}
  =\frac{p(\theta)}{p(\theta')}.
  \label{eq:survivor-ratio}
\end{equation}

The observed record removes contradicted models while preserving the prior odds among
survivors.

\section{Realisation on ARC-AGI-3}
\label{sec:realisation}

\subsection{Environment and scoring}
\label{sec:game}

\begin{definition}[ARC-AGI-3 game]
        We model an ARC-AGI-3 level as a deterministic, goal-directed POMDP
of the form in Section~\ref{sec:formulation},
\begin{equation}
  \cM=\langle\cS,\cA,\cO,T,\rho,G\rangle,
  \qquad s_{t+1}=T(s_t,a_t),\quad o_t=\rho(s_t).
  \label{eq:game}
\end{equation}
        The visual component of each observation is a $64\times64$ colour grid. In the final level of \texttt{ls20}, for example,
fog of war masks regions outside the agent's local view, so the current frame does not reveal
the full environment state. Actions are drawn from the supplied key/click interface, with
initially unknown semantics, and $G$ contains the states that complete a level. The agent
observes this interface and its own interaction history, but not the environment source or
a language description of the game.
\end{definition}

For a sequence of levels, completion is prioritised first and total interaction cost is
compared among runs completing the same set of levels.
For a game with $L$ levels, let $h_\ell$ and $a_\ell$ denote the human and agent
action counts on level $\ell$, and let $C$ be the set of levels completed by the agent.
ARC-AGI-3 scores action efficiency by
\begin{equation}
  \begin{aligned}
    e_\ell
      &=\min\{1.15,(h_\ell/a_\ell)^2\},\\
    \mathrm{RHAE}
      &=100\,\min\!\left\{
        \frac{\sum_{\ell\in C}\ell}{\sum_{\ell=1}^{L}\ell},
        \frac{\sum_{\ell=1}^{L}\ell e_\ell}{\sum_{\ell=1}^{L}\ell}
      \right\},
  \end{aligned}
  \label{eq:rhae}
\end{equation}
with $a_\ell=\infty$ for an uncleared level. Model construction, local execution, and replay do not add
to the action count. All actions sent to the environment count, including resets. Later levels can
introduce new mechanics, so the agent must be able to revise its rulebook and executable throughout the game.

\subsection{Rulebook and executable artefacts}
\label{sec:arc-artifacts}

\begin{figure}[p]
  \centering
  \includegraphics[width=\linewidth]{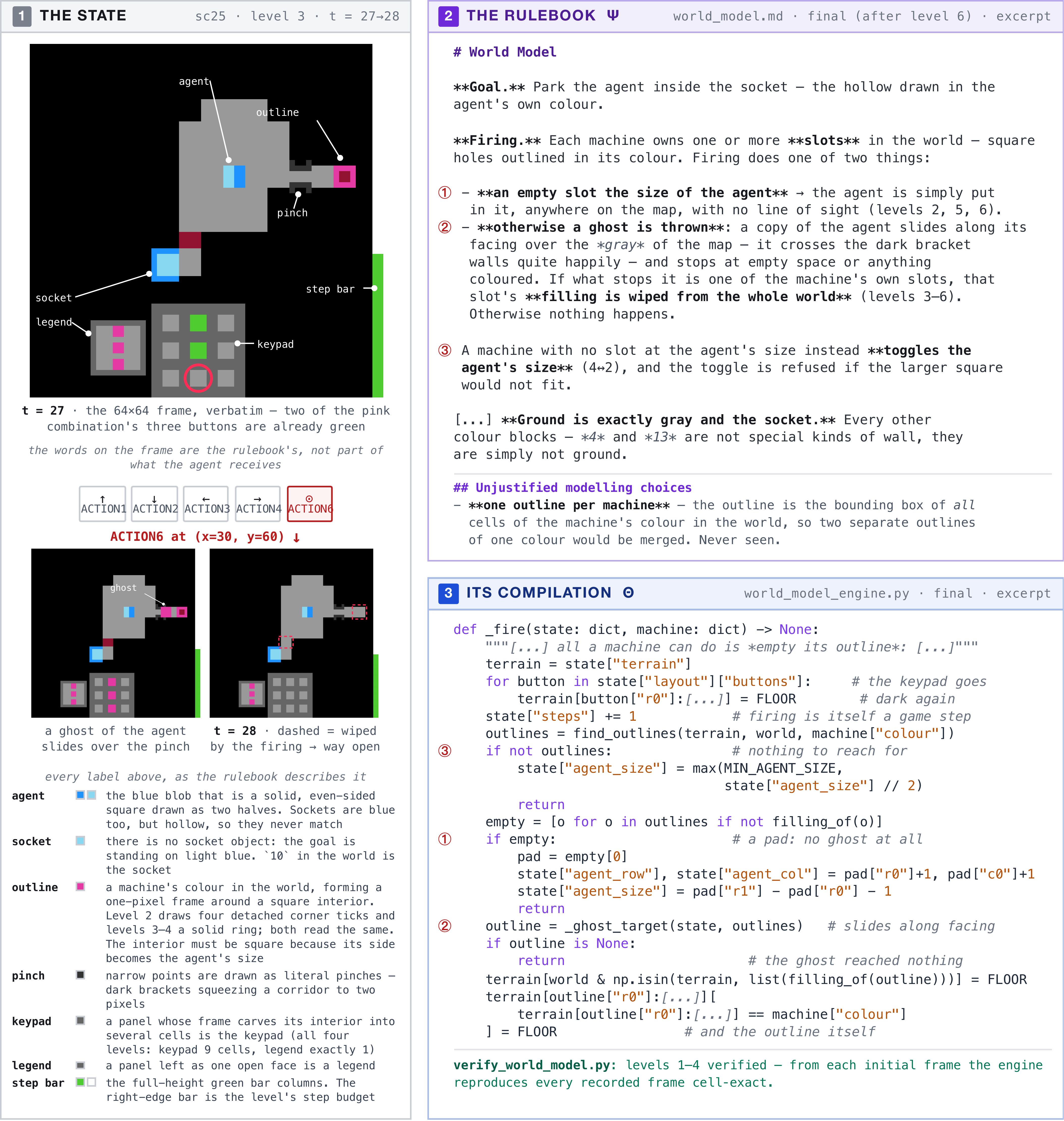}
  \caption{An ARC-AGI-3 realisation of a rulebook and its executable. Left: an observed transition
  from game \texttt{sc25}. Upper right: the final rulebook in \texttt{world\_model.md}, whose names
  are also used to annotate the frame. Lower right: an excerpt from
  \texttt{world\_model\_engine.py} implementing the numbered rules. The executable is checked by
  cell-exact replay before it is used for planning.}
  \label{fig:example}
\end{figure}

We realise the abstract rulebook $\psi_t$ as a Markdown file named
\texttt{world\_model.md}. It records the agent's current account of objects, action semantics,
dynamics, goals, and unresolved alternatives in language. The executable $\theta_t$ includes the Python
module \texttt{world\_model\_engine.py} for internal state transitions, together with routines for
state reconstruction, rendering, and checking goal completion. It predicts the
complete next grid and whether a level has been completed. Figure~\ref{fig:example}
illustrates these artefacts through a concrete example, linking an observed transition to the
corresponding rulebook entries and executable code.

The frozen LLM implements reflection, rule revision, and compilation by inspecting the accumulated
frames and editing the rulebook and executable source. Its fidelity check compares the source with
the rulebook. For replay, a new trajectory starts with the observation returned after each reset or
level start. The initial observation $o_0$ includes the starting grid and the level information
returned by the interface. The verifier initialises the model with $\iota_\theta(o_0)$, then replays the recorded actions. Rendered predictions
must match all $64\times64$ cells of every subsequent frame within the trajectory before the
executable is made available to the planner.
The planner searches the predicted dynamics from the reconstructed current model state for
a route to the inferred goal and executes the resulting action sequence one action at a time.
Execution stops at the first prediction mismatch so that the new counterexample can be incorporated
before planning continues.

\subsection{Continual operation}
\label{sec:arc-operation}

\begin{figure}[!t]
  \centering
  \includegraphics[width=\linewidth]{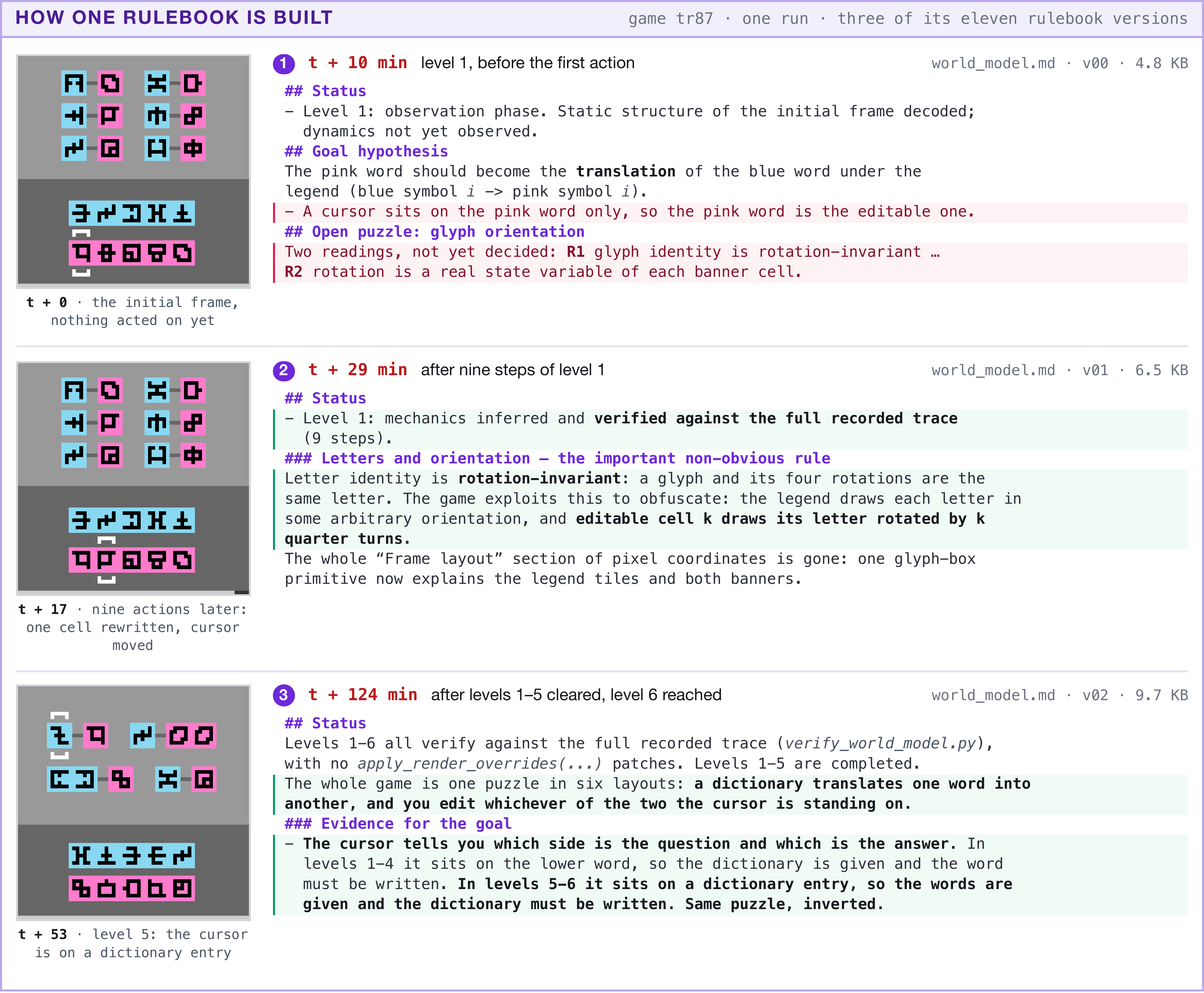}
  \caption{Three of the eleven rulebook versions from one run on game \texttt{tr87}, each paired
  with the observation that prompted the revision. (1) Before the first action, the rulebook records
  two competing hypotheses about glyph orientation. (2) Interaction establishes rotation-invariant
  letter identity and replaces pixel coordinates with a glyph-box rule. (3) Evidence from later
  levels generalises and reverses the cursor rule. Red marks a claim retracted by a later version and
  green marks the statement that resolves it.}
  \label{fig:evolution}
\end{figure}

The general update in Section~\ref{sec:loop} runs throughout each game rather than only during
initial model construction. The harness retains the full interaction record $H_t$, including all trajectories, and invokes reflection whenever new evidence is available.
The learned rulebook and executable persist across attempts and levels, allowing rules inferred
earlier to guide prediction and planning as further mechanics are encountered. A rule-level change is written to the Markdown
rulebook before the Python executable is refined. An implementation-only diagnosis leaves the
rulebook intact and changes the Python module. Figure~\ref{fig:evolution} provides a detailed example
of rulebook revision across three versions, showing how new observations resolve an initial
ambiguity and later overturn an earlier rule.

\subsection{Git memory}
\label{sec:git}

\codemodel stores the agent's model history in a Git repository. The rulebook and the modules
compiled from it live as files in a single version-controlled workspace. After each iteration, the
controller commits the workspace and tags the commit with the iteration number. Without version
control, each revision destructively overwrites the rulebook or implementation it replaces.
Here, every $\psi$ and every $\theta$ the run has held remains addressable, together with when it was
written. The repository therefore records the agent's learning trajectory, not just a snapshot of its
latest state.

Crucially, this history is available to the agent itself. It can inspect earlier iterations, diff two
versions of the rulebook or of a module, and recover their contents at any previous commit. The agent
can therefore recall hypotheses it has already tested, examine the consequences of a past revision,
or restore an executable that a later edit broke.

\section{Experiments}
\label{sec:exp}

\begin{figure}[!htb]
  \centering
  \includegraphics[width=\linewidth]{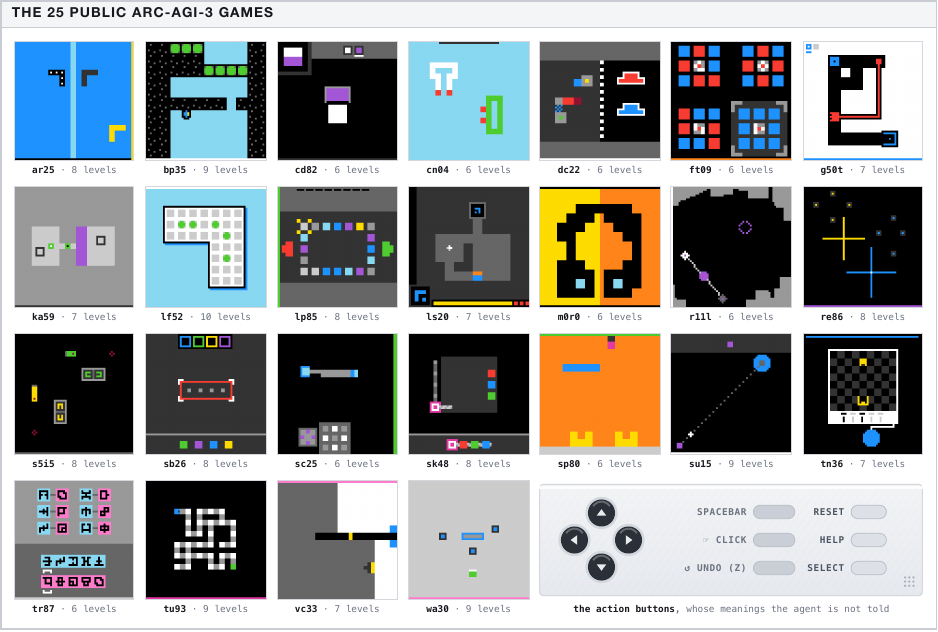}
  \caption{\textbf{The $\numgames$ public ARC-AGI-3 games.}}
  \label{fig:games}
\end{figure}

\paragraph{Benchmark and setup.}
We evaluate on all $\numgames$ public ARC-AGI-3 games~\citep{arcagi3}, shown in
Figure~\ref{fig:games}. The task and its scoring rule are defined in Section~\ref{sec:game}. For each
run, we count every action sent to the environment, including actions used in failed attempts and
after the reset that follows. We report the RHAE of Eq.~\eqref{eq:rhae}, for which higher is better
and $100$ is the ceiling. The agent observes only the action interface and cannot inspect the
environment source.

We build our experimental harness using the baseline1 codebase~\citep{rodionov2026executable} as a
starting point and run it in Claude Code with Claude Opus~5 as the backbone model and reasoning
effort set to extra-high.
Every baseline is an entry on the ARC-AGI-3 community leaderboard that links to a paper:
\mbox{baseline1}~\citep{rodionov2026executable}, NOOA~\citep{nvidialabs2026nooa},
OPINE-World~\citep{opineworld}, DreamTeam~\citep{sarafian2026workspace} and Continual
Harness~\citep{continualharness}. Every baseline number, including per-game RHAE, actions and levels
cleared, comes from that system's official ARC Prize \mbox{scorecard}, as does the human action baseline,
so all columns are counted the same way.

\paragraph{Main result.}
Table~\ref{tab:results} reports every game. \arcworld reaches the $100$ ceiling on all
$\numceiling$ games, yielding a \textbf{mean RHAE of \ourrhae}. A ceiling score requires every
level to be cleared and the level-weighted mean of $e_\ell$ in Eq.~\eqref{eq:rhae} to be at least $1$.
Table~\ref{tab:actions} further shows that the total action count for every game is below its human
baseline. Among the systems compared in Table~\ref{tab:results}, the closest is baseline1 at
$\baseonerhae$, which also clears every game; NOOA reaches $\nooarhae$ having cleared $19$ of
the $\numgames$ games, OPINE-World $\opinerhae$ with $20$, DreamTeam $\dtrhae$ with $6$, and
Continual Harness $\chrhae$ with $3$.

\begin{table}[!htb]
\centering
\footnotesize
\caption{\textbf{Per-game RHAE on the $\numgames$ public ARC-AGI-3 games.} $L$ is the number of
levels in the game. RHAE is the level-weighted human-relative efficiency of Eq.~\eqref{eq:rhae}.
All baseline numbers come from the official ARC Prize scorecards.}
\label{tab:results}
\renewcommand{\arraystretch}{1.05}
\setlength{\tabcolsep}{4pt}
\begin{tabular}{@{}l r rrrrr >{\columncolor{oursbg}}r@{}}
\toprule
\shortstack{\\Game} & \shortstack{\\$L$} & \shortstack{Continual\\Harness} & \shortstack{Dream-\\Team} & \shortstack{OPINE-\\World} & \shortstack{\\NOOA} & \shortstack{\\baseline1} & \shortstack{\\\textbf{Ours}} \\
\midrule
\texttt{ar25} & 8 & 17.4 & 100.0 & 100.0 & 100.0 & 100.0 & \textbf{100} \\
\texttt{bp35} & 9 & 1.1 & 0.8 & 2.5 & 46.2 & 100.0 & \textbf{100} \\
\texttt{cd82} & 6 & 0.0 & 79.0 & 100.0 & 100.0 & 100.0 & \textbf{100} \\
\texttt{cn04} & 6 & 46.4 & 0.0 & 100.0 & 100.0 & 100.0 & \textbf{100} \\
\texttt{dc22} & 6 & 18.2 & 0.0 & 82.4 & 64.7 & 100.0 & \textbf{100} \\
\addlinespace[1.5pt]
\texttt{ft09} & 6 & 70.1 & 100.0 & 100.0 & 100.0 & 100.0 & \textbf{100} \\
\texttt{g50t} & 7 & 10.7 & 2.9 & 61.6 & 65.3 & 100.0 & \textbf{100} \\
\texttt{ka59} & 7 & 38.6 & 59.7 & 62.5 & 100.0 & 100.0 & \textbf{100} \\
\texttt{lf52} & 10 & 0.2 & 10.9 & 4.2 & 27.3 & 100.0 & \textbf{100} \\
\texttt{lp85} & 8 & 100.0 & 100.0 & 100.0 & 100.0 & 100.0 & \textbf{100} \\
\addlinespace[1.5pt]
\texttt{ls20} & 7 & 3.6 & 14.7 & 71.2 & 37.4 & 100.0 & \textbf{100} \\
\texttt{m0r0} & 6 & 66.8 & 47.6 & 100.0 & 100.0 & 100.0 & \textbf{100} \\
\texttt{r11l} & 6 & 4.8 & 47.6 & 100.0 & 100.0 & 100.0 & \textbf{100} \\
\texttt{re86} & 8 & 16.7 & 58.3 & 100.0 & 100.0 & 100.0 & \textbf{100} \\
\texttt{s5i5} & 8 & 0.0 & 38.2 & 27.1 & 100.0 & 100.0 & \textbf{100} \\
\addlinespace[1.5pt]
\texttt{sb26} & 8 & 0.0 & 94.0 & 89.8 & 100.0 & 100.0 & \textbf{100} \\
\texttt{sc25} & 6 & 0.0 & 36.1 & 84.0 & 83.5 & 84.2 & \textbf{100} \\
\texttt{sk48} & 8 & 1.1 & 6.9 & 21.3 & 72.5 & 100.0 & \textbf{100} \\
\texttt{sp80} & 6 & 47.6 & 0.1 & 100.0 & 79.2 & 100.0 & \textbf{100} \\
\texttt{su15} & 9 & 0.0 & 2.2 & 92.0 & 97.0 & 100.0 & \textbf{100} \\
\addlinespace[1.5pt]
\texttt{tn36} & 7 & 0.0 & 5.0 & 68.4 & 100.0 & 90.1 & \textbf{100} \\
\texttt{tr87} & 6 & 0.0 & 12.6 & 100.0 & 100.0 & 100.0 & \textbf{100} \\
\texttt{tu93} & 9 & 61.4 & 100.0 & 100.0 & 93.4 & 100.0 & \textbf{100} \\
\texttt{vc33} & 7 & 8.9 & 21.4 & 92.1 & 99.6 & 100.0 & \textbf{100} \\
\texttt{wa30} & 9 & 0.0 & 13.3 & 100.0 & 62.2 & 100.0 & \textbf{100} \\
\midrule
\textbf{Mean, all 25} & \textbf{183} & 20.5 & 38.1 & 78.4 & 85.1 & 99.0 & \textbf{100.0} \\
\textit{Games cleared} &  & 3 & 6 & 20 & 19 & 25 & 25 \\
\textit{Levels cleared} &  & 64 & 98 & 160 & 170 & 183 & 183 \\
\bottomrule
\end{tabular}

\end{table}

Table~\ref{tab:actions} compares action counts. Our agent clears all
\numgames{} games in \ouractions{} actions, \actionratio{} of the human baseline
(\humanactions{}). Per-game ratios range from $0.20$ on \texttt{m0r0} to
$0.67$ on \texttt{lf52} (median $0.42$). Among the systems in Table~\ref{tab:actions},
baseline1 is the only other system to finish every game, using \baseactions{} actions
($0.49$ of the human budget).
The other four leave games unfinished: OPINE-World leaves $5$, NOOA $6$, DreamTeam
$19$ and Continual Harness $22$. Their totals (\opineactions{}, \nooaactions{},
\dtactions{} and \chactions{}) therefore reflect incomplete runs rather than solution costs.

\begin{table}[!htb]
\centering
\footnotesize
\caption{\textbf{Per-game action counts.} Results are from the same runs and scorecards as
Table~\ref{tab:results}. The bar shows our count as a fraction of the human count, so a full track
would indicate parity. A dagger marks a system that left at least one level of the game uncleared.
Its count is therefore what the run spent rather than what solving the game requires. Totals marked
with a dagger are not comparable as solution costs.}
\label{tab:actions}
\renewcommand{\arraystretch}{1.05}
\setlength{\tabcolsep}{4pt}
\begin{tabular}{@{}l r@{\hskip 7pt} rrrrr >{\columncolor{oursbg}}r r@{\,}l@{}}
\toprule
\shortstack{\\Game} & \shortstack{\\Human} & \shortstack{Continual\\Harness} & \shortstack{Dream-\\Team} & \shortstack{OPINE-\\World} & \shortstack{\\NOOA} & \shortstack{\\baseline1} & \shortstack{\\\textbf{Ours}} & \multicolumn{2}{c}{\shortstack{\\Ours\,/\,Human}} \\
\midrule
\texttt{ar25} & 748 & 687$^{\dagger}$ & 479 & 381 & 274 & 271 & \textbf{253} & 0.34 & \ratiobar{0.338} \\
\texttt{bp35} & 651 & 274$^{\dagger}$ & 647$^{\dagger}$ & 512$^{\dagger}$ & 623$^{\dagger}$ & 532 & \textbf{429} & 0.66 & \ratiobar{0.659} \\
\texttt{cd82} & 171 & 275$^{\dagger}$ & 282 & 161 & 115 & 93 & \textbf{82} & 0.48 & \ratiobar{0.480} \\
\texttt{cn04} & 789 & 1{,}792$^{\dagger}$ & 444$^{\dagger}$ & 264 & 441 & 201 & \textbf{201} & 0.25 & \ratiobar{0.255} \\
\texttt{dc22} & 1{,}228 & 1{,}097$^{\dagger}$ & 355$^{\dagger}$ & 1{,}485 & 841$^{\dagger}$ & 898 & \textbf{643} & 0.52 & \ratiobar{0.524} \\
\addlinespace[1.5pt]
\texttt{ft09} & 208 & 270 & 82 & 112 & 84 & 85 & \textbf{75} & 0.36 & \ratiobar{0.361} \\
\texttt{g50t} & 879 & 1{,}046$^{\dagger}$ & 395$^{\dagger}$ & 758 & 728 & 363 & \textbf{307} & 0.35 & \ratiobar{0.349} \\
\texttt{ka59} & 730 & 796$^{\dagger}$ & 510$^{\dagger}$ & 1{,}077$^{\dagger}$ & 449 & 362 & \textbf{341} & 0.47 & \ratiobar{0.467} \\
\texttt{lf52} & 1{,}339 & 950$^{\dagger}$ & 230$^{\dagger}$ & 594$^{\dagger}$ & 995$^{\dagger}$ & 767 & \textbf{894} & 0.67 & \ratiobar{0.668} \\
\texttt{lp85} & 388 & 344 & 120 & 110 & 114 & 90 & \textbf{95} & 0.24 & \ratiobar{0.245} \\
\addlinespace[1.5pt]
\texttt{ls20} & 776 & 634$^{\dagger}$ & 556$^{\dagger}$ & 959 & 1{,}130$^{\dagger}$ & 465 & \textbf{473} & 0.61 & \ratiobar{0.610} \\
\texttt{m0r0} & 1{,}107 & 1{,}705 & 367$^{\dagger}$ & 260 & 241 & 215 & \textbf{220} & 0.20 & \ratiobar{0.199} \\
\texttt{r11l} & 233 & 178$^{\dagger}$ & 263$^{\dagger}$ & 128 & 87 & 97 & \textbf{75} & 0.32 & \ratiobar{0.322} \\
\texttt{re86} & 1{,}255 & 716$^{\dagger}$ & 540$^{\dagger}$ & 851 & 803 & 559 & \textbf{625} & 0.50 & \ratiobar{0.498} \\
\texttt{s5i5} & 638 & 100$^{\dagger}$ & 365$^{\dagger}$ & 639$^{\dagger}$ & 317 & 283 & \textbf{261} & 0.41 & \ratiobar{0.409} \\
\addlinespace[1.5pt]
\texttt{sb26} & 213 & 100$^{\dagger}$ & 210 & 215 & 128 & 140 & \textbf{124} & 0.58 & \ratiobar{0.582} \\
\texttt{sc25} & 350 & 180$^{\dagger}$ & 486$^{\dagger}$ & 257 & 488 & 619 & \textbf{173} & 0.49 & \ratiobar{0.494} \\
\texttt{sk48} & 1{,}070 & 1{,}215$^{\dagger}$ & 331$^{\dagger}$ & 596$^{\dagger}$ & 1{,}127$^{\dagger}$ & 419 & \textbf{407} & 0.38 & \ratiobar{0.380} \\
\texttt{sp80} & 518 & 587$^{\dagger}$ & 448$^{\dagger}$ & 370 & 444 & 193 & \textbf{144} & 0.28 & \ratiobar{0.278} \\
\texttt{su15} & 361 & 112$^{\dagger}$ & 558$^{\dagger}$ & 335 & 200 & 117 & \textbf{104} & 0.29 & \ratiobar{0.288} \\
\addlinespace[1.5pt]
\texttt{tn36} & 317 & 163$^{\dagger}$ & 509$^{\dagger}$ & 418 & 137 & 292 & \textbf{133} & 0.42 & \ratiobar{0.420} \\
\texttt{tr87} & 414 & 272$^{\dagger}$ & 528$^{\dagger}$ & 212 & 438 & 147 & \textbf{148} & 0.36 & \ratiobar{0.357} \\
\texttt{tu93} & 462 & 456$^{\dagger}$ & 260 & 272 & 229 & 201 & \textbf{208} & 0.45 & \ratiobar{0.450} \\
\texttt{vc33} & 447 & 254$^{\dagger}$ & 401$^{\dagger}$ & 427 & 263 & 293 & \textbf{204} & 0.46 & \ratiobar{0.456} \\
\texttt{wa30} & 1{,}843 & 355$^{\dagger}$ & 423$^{\dagger}$ & 1{,}466 & 1{,}001$^{\dagger}$ & 645 & \textbf{899} & 0.49 & \ratiobar{0.488} \\
\midrule
\textbf{Total, all 25} & 17{,}135 & 14{,}558$^{\dagger}$ & 9{,}789$^{\dagger}$ & 12{,}859$^{\dagger}$ & 11{,}697$^{\dagger}$ & 8{,}347 & \textbf{7{,}518} & \textbf{0.44} & \ratiobar{0.439} \\
\bottomrule
\end{tabular}

\end{table}

\paragraph{Ablations.}
To isolate the contribution of the natural-language rulebook, we remove it from the update loop. This
leaves a code world model while keeping the harness, the replay verifier, the plan executor, the
backbone and the reasoning effort byte-identical. Given the difficulty of ARC-AGI-3 tasks, each
experimental run with Claude Code incurs substantial time and monetary cost. We therefore select one
game from each of three difficulty bands defined by human actions per level. This normalisation
separates the interaction difficulty of a level from the number of levels in a game. The selected
games are \texttt{ft09} at $35$ actions per level for the easy band,
\texttt{ka59} at $104$ for the medium band and \texttt{cn04} at $132$ for the hard band. Every run
clears every level and reaches RHAE~$100$, so Table~\ref{tab:mdabl} compares the cost of convergence.
Actions count scored environment interactions, while agent turns count agent iterations to completion
and are unaffected by wall-clock delays.

In the reported runs, the rulebook variant uses fewer actions and fewer agent turns in every difficulty
band. Across the three games, it uses $617$ rather than $677$ actions, a reduction of $9\%$, and $678$
rather than $830$ agent turns, a reduction of $18\%$. The reduction in actions is consistent across all
three games, while the largest reductions in agent turns occur on \texttt{ft09} and \texttt{cn04}.
With task completion held constant, these results suggest that the persistent rulebook improves
convergence to an effective executable model.

\begin{table}[!htb]
\centering
\footnotesize
\caption{\textbf{Ablating the rulebook, one game per difficulty band.} Difficulty is the human
action baseline divided by the level count.}
\label{tab:mdabl}
\renewcommand{\arraystretch}{1.05}
\setlength{\tabcolsep}{5pt}
\begin{tabular}{@{}l l r@{\hskip 12pt} >{\columncolor{oursbg}}r r@{\hskip 12pt} >{\columncolor{oursbg}}r r@{}}
\toprule
 & & & \multicolumn{2}{c}{Actions} & \multicolumn{2}{c}{Agent turns} \\
\cmidrule(lr){4-5}\cmidrule(lr){6-7}
\shortstack{\\Game} & \shortstack{\\Band} & \shortstack{Human actions\\per level} &
\shortstack{with\\\textbf{rulebook}} & \shortstack{without\\rulebook} &
\shortstack{with\\\textbf{rulebook}} & \shortstack{without\\rulebook} \\
\midrule
\texttt{ft09} & Easy   & 35  & \textbf{75}  & 79  & \textbf{132}   & 174 \\
\texttt{ka59} & Medium & 104 & \textbf{341} & 382 & \textbf{298} & 310 \\
\texttt{cn04} & Hard   & 132 & \textbf{201} & 216 & \textbf{248}  & 346 \\
\bottomrule
\end{tabular}

\end{table}

\paragraph{Population.}
Section~\ref{sec:map} generalises \codemodel from one rulebook--executable pair to a population of
$N$ admissible pairs that share the interaction history.
Under this population setting, the system samples one pair for each planning episode.
This shared-history population is designed to reduce
run-to-run variance in interactive model-based control by preventing a single stochastic initial model
from determining the entire trajectory. The resulting interaction is used to evaluate all pairs,
and any pair whose prediction conflicts with the observation is repaired. The main experiments use
the single-pair setting ($N=1$). To examine the population extension, we select \texttt{wa30}, which has
the largest human action baseline among the $25$ public games, and run it at $N=2$. We hold the
backbone, the reasoning effort and the harness fixed at their single-model settings, so only the
number of retained pairs changes. During execution, a communication mechanism keeps both members
synchronised. It informs them which member supplied the active plan, which actions were sent to the
environment and what real transitions followed, so both members update from the same evidence.

Table~\ref{tab:population} reports the result. Both arms clear all nine levels and score RHAE~$100$
when replayed against the live server. Compared with the verified single-model trajectory, the population
reduces the total from $899$ to $597$ scored actions and matches or improves per-level action
efficiency on eight of the nine levels. The two world models begin with different hypotheses and
converge as the shared interaction
history resolves their disagreements. Together, these results suggest that maintaining a population
can make world-model-guided exploration more action-efficient and less sensitive to any single
initial hypothesis.

\begin{table}[!htb]
\centering
\footnotesize
\caption{\textbf{Population extension, level by level on \texttt{wa30}.}}
\label{tab:population}
\renewcommand{\arraystretch}{1.05}
\setlength{\tabcolsep}{5pt}
\begin{tabular}{@{}l r r@{\hskip 12pt} >{\columncolor{oursbg}}r r@{}}
\toprule
 & & \multicolumn{2}{c}{Actions} & \\
\cmidrule(lr){3-4}
\shortstack{\\Level} & \shortstack{Human\\baseline} &
\shortstack{$N{=}1$} & \shortstack{$\mathbf{N{=}2}$\\\textbf{population}} &
\shortstack{$N{=}2$ vs\\$N{=}1$} \\
\midrule
L1 & 71   & 42  & 26  & $0.62\times$ \\
L2 & 119  & 60  & 58  & $0.97\times$ \\
L3 & 183  & 98  & 76  & $0.78\times$ \\
L4 & 98   & 84  & 50  & $0.60\times$ \\
L5 & 368  & 245 & 106 & $0.43\times$ \\
L6 & 68   & 54  & 46  & $0.85\times$ \\
L7 & 79   & 49  & 36  & $0.73\times$ \\
L8 & 442  & 130 & 138 & $1.06\times$ \\
L9 & 415  & 137 & 61  & $0.45\times$ \\
\midrule
Total & 1843 & 899 & \textbf{597} & $\mathbf{0.66\times}$ \\
RHAE  &      & 100 & \textbf{100} & \\
\bottomrule
\end{tabular}

\end{table}

\section{Case Study on Atari Pong}
\label{sec:atari}

In ARC-AGI-3, the agent interacts with an unknown deterministic environment, where each level
presents a discrete goal and performance is measured by action efficiency. The executable world
model supports a planner that searches for a finite action sequence to complete the current level.
This raises a broader question: can the same rulebook-based learning mechanism also support
feedback control, where the agent repeatedly updates its decisions from new observations?
We investigate this question in Atari Pong~\citep{bellemare2013ale}, a two-player adversarial
setting in which the agent must model both the physical dynamics and the opponent's behaviour.
The agent controls one paddle and must anticipate how the ball and the opposing paddle will move.

\begin{figure}[!htb]
  \centering
  \includegraphics[width=\linewidth]{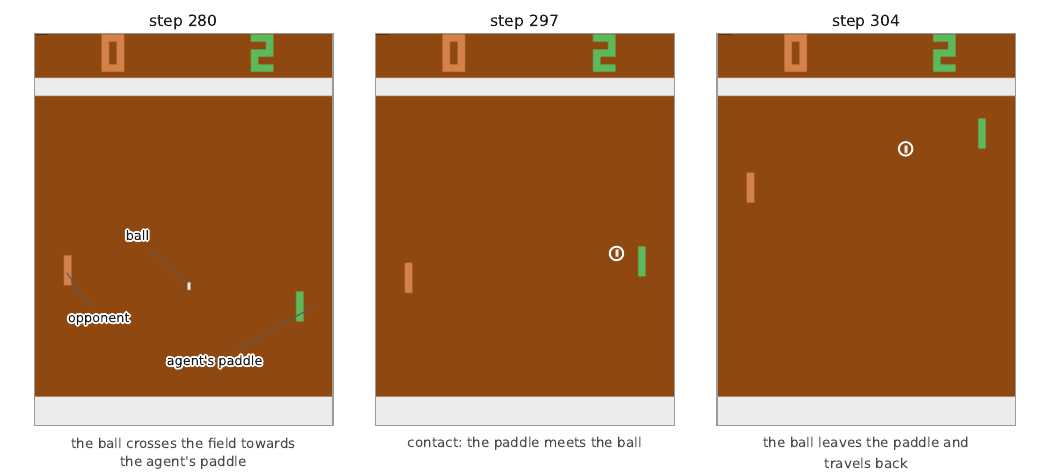}
  \caption{\textbf{How the Pong game works.} The agent controls the right-hand paddle and the
  opponent controls the left-hand paddle. Each player moves its paddle up or down to intercept
  the ball and return it towards the other side, scoring a point when the opponent fails to
  return it. The three frames illustrate one return: the ball approaches the agent's paddle,
  makes contact, and bounces back towards the opponent.}
  \label{fig:pong}
\end{figure}

\paragraph{Setup.}
The agent learns Pong dynamics through interaction, revising its rulebook and executable
as new evidence arrives.
\Cref{fig:pong} shows the game.
Emulation uses the Arcade Learning Environment (ALE)~\citep{bellemare2013ale} with frame-skip $4$.
For Pong, we replace the level-solving planner with a feedback policy that uses the learned
executable world model to predict the consequences of candidate actions and selects the next
action as new observations arrive.

Validation in this feedback-control setting combines replay diagnostics with evaluation
of the learned controller. Discrepancies between predicted and observed
ball trajectories, paddle motion, and bounce outcomes guide revisions to the rulebook
and code. Full-frame replay retains residual rendering and paddle-motion errors.
Before each action, the controller reconstructs its current state from recent frames and
actions, fitting latent quantities such as ball velocity and the opponent's movement phase
under the learned dynamics. This updated state supplies the starting point for predicting
action consequences.

During learning, we evaluate checkpoints of the learned controller in separate game episodes.
The controller remains fixed during evaluation and requires no LLM calls. Exploration and
revision continue until it achieves a $21\!:\!0$ win, at which point learning stops.

\enlargethispage{\baselineskip}
\begin{table}[!t]
\centering
\footnotesize
\caption{\textbf{Pong score.} Following the evaluation metric used in reinforcement learning
experiments, episode score is defined as the agent's points minus the opponent's, ranging from
$-21$ to $+21$. Our score is averaged over the three episodes shown in \Cref{fig:pong-curves},
while the published baselines follow their respective evaluation protocols.}
\label{tab:atari}
\renewcommand{\arraystretch}{1.05}
\setlength{\tabcolsep}{4pt}
\begin{tabular}{@{}l l r r@{}}
\toprule
Method & Approach & Score & Learning cost \\
& & & (frames) \\
\midrule
GDI~\citep{fan2021gdi}
  & Model-free RL, data-distribution iteration
  & $21$    & $2.0\times10^{8}$ \\
LASER~\citep{schmitt2020laser}
  & Model-free RL, actor--critic with replay
  & $21$    & $2.0\times10^{8}$ \\
IMPALA~\citep{espeholt2018impala}
  & Model-free RL, distributed actor--critic
  & $20.98$ & $2.0\times10^{8}$ \\
Rainbow~\citep{hessel2018rainbow}
  & Model-free RL, distributional DQN with replay
  & $20.9$  & $2.0\times10^{8}$ \\
EfficientZero~\citep{ye2021efficientzero}
  & Model-based RL, latent dynamics and MCTS
  & $20.1$ & $4.0\times10^{5}$ \\
EfficientZero V2~\citep{wang2024efficientzerov2}
  & Model-based RL, latent dynamics and tree search
  & $20.8$ & $4.0\times10^{5}$ \\
Human average
  & Human play & $14.6$ & --- \\
Same LLM, no world model
  & One LLM call per frame, memoryless & $-19$ & $0$ \\
Random
  & Uniform over the action set & $-20.7$ & --- \\
\midrule
\rowcolor{oursbg}
\textbf{\codemodel}
  & \textbf{Language rulebook compiled to code}
  & $\bm{21}$ & $\bm{9.5\times10^{3}}$ \\
\bottomrule
\end{tabular}

\end{table}

\begin{figure}[!t]
  \centering
  \includegraphics[width=\linewidth]{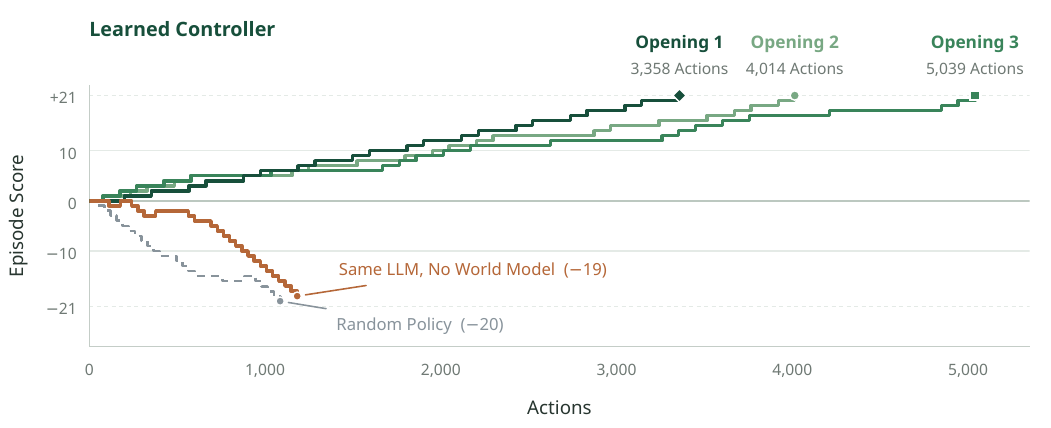}
  \caption{\textbf{Pong episode scores.} The three green curves show the learned controller
  under three different openings. Each reaches $+21$ without conceding a point. The same LLM
  without a world model finishes with a score of $-19$, and the random-policy episode shown here
  finishes at $-20$. The horizontal axis counts actions taken within each episode.}
  \label{fig:pong-curves}
\end{figure}

\begin{figure}[!t]
  \centering
  \includegraphics[width=\linewidth]{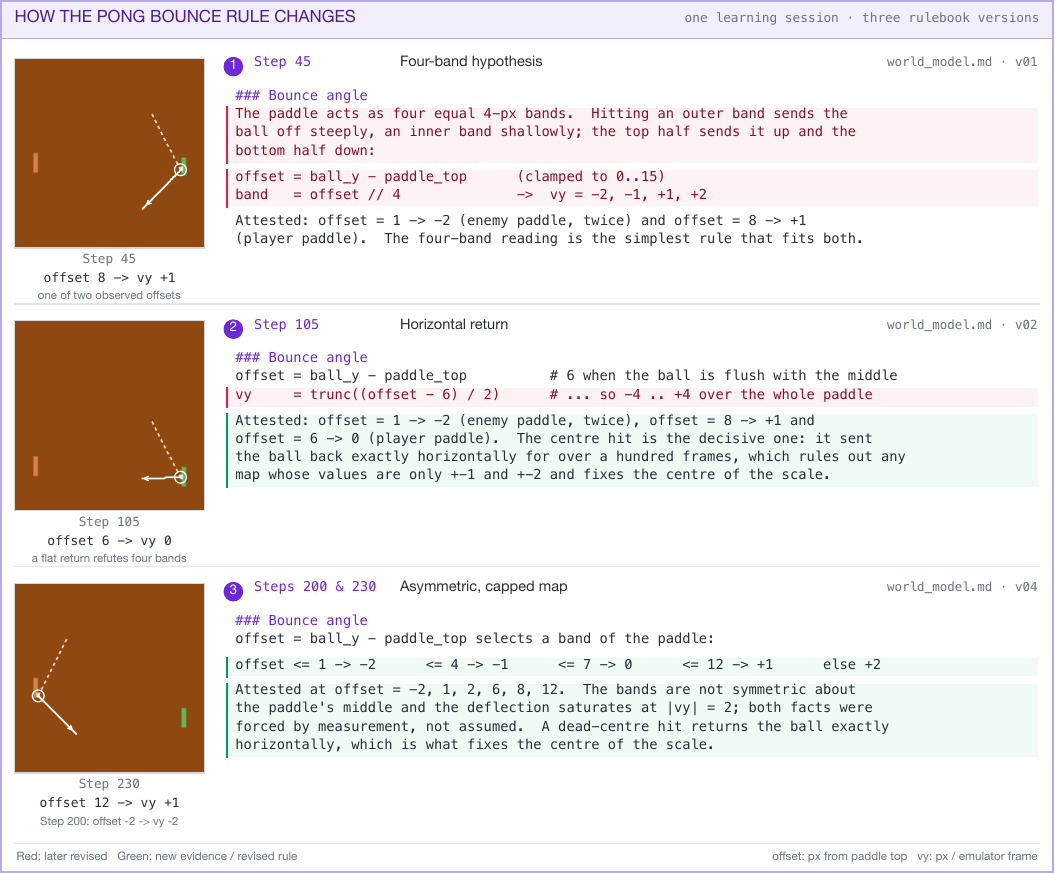}

  \caption{Three rulebook versions from one Pong learning session show how observed ball returns drive two
  revisions of the bounce rule shared by both paddles.
  (1) The initial four-band hypothesis excludes horizontal returns.
  (2) A horizontal return from a centre hit refutes this hypothesis and prompts a
  centred-deflection formula that permits $v_y=0$.
  (3) Further contacts reveal that the formula overestimates vertical return speeds,
  leading to an asymmetric band map with $|v_y|\leq2$.
  Contact offset is the ball's top row minus the paddle's top row; $v_y$ denotes vertical
  velocity in pixels per emulator frame. Red marks claims revised later, and green marks
  new evidence or revised rules. Dashed white trails show recorded ball positions before
  contact, while solid white trails show positions after contact; step labels identify
  the supporting observations.}
  \label{fig:pong-evolution}
\end{figure}

\paragraph{Baselines.}
We compare our method with three types of baselines: a direct LLM agent with chain-of-thought reasoning,
model-free reinforcement learning (RL) methods, and model-based RL methods (\Cref{tab:atari}).

The direct LLM agent
uses the same  backbone, environment, and action interface but has no
rulebook or executable world model. This baseline selects each action from the two most recent
observations, with no persistent memory between calls. We also evaluate a uniformly random policy as a reference.

Among the model-free RL baselines,
GDI~\citep{fan2021gdi} optimises both policy learning and the distribution of training data.
LASER~\citep{schmitt2020laser} is an off-policy actor--critic method with shared experience
replay, while IMPALA~\citep{espeholt2018impala} uses distributed actor--critic learning with
V-trace correction. Rainbow~\citep{hessel2018rainbow} combines several improvements to the
deep Q-network (DQN) algorithm. Their reported scores, alongside human and random-policy
scores, are taken from \citet{fan2021atarireview}.
The model-based RL baselines, EfficientZero~\citep{ye2021efficientzero} and
EfficientZero V2~\citep{wang2024efficientzerov2}, learn latent dynamics and use tree search
for action selection, with a training budget of $100{,}000$ actions, equivalent to $400{,}000$ emulator frames.

\paragraph{Result.}
We evaluate the learned controller in three episodes, each with a different opening. It wins
$21\!:\!0$ in all three, achieving the maximum episode score of $21$.
Openings~1, 2 and~3 require $3{,}358$, $4{,}014$ and $5{,}039$ actions, respectively.
The LLM is used only during learning. During evaluation, the controller is held fixed,
as is a trained reinforcement learning policy. \Cref{tab:atari} reports episode scores and
learning costs for our controller and the baselines. Learning cost is measured in emulator frames.
\Cref{fig:pong-curves} shows the score trajectories
of the three controller episodes alongside those of the no-world-model and random-policy baselines.

\Needspace{3\baselineskip}
\begingroup
\widowpenalty=10000
Our method achieves the maximum episode score of $21$ at a learning cost of $9{,}504$ emulator frames.
The reported learning costs of the model-free and model-based baselines in \Cref{tab:atari}
are approximately $2.1\times10^4$ and $42$ times this cost, respectively.
The rulebook and compiled code support learning from limited interaction by turning the
frozen LLM's hypotheses into an explicit world model that can be revised and reused for
control. Drawing on pretrained knowledge, the LLM proposes objects, state variables, and
dynamical relationships, which the rulebook records for testing against new observations.
\Cref{fig:pong-evolution} illustrates this process: a centre hit refutes
the initial four-band bounce hypothesis and prompts a centred-deflection formula; further
contacts lead to an asymmetric band map. Each revision updates the bounce rule for both
paddles and changes its predictions at unobserved contact offsets.
Compilation makes each revised rule available to the controller as an executable function.
Combined with the learned ball and opponent dynamics, this function supports comparison of
candidate return trajectories without testing each one in the environment.
The model need not recover the dynamics of every possible state; it must be sufficiently
accurate for the decisions required for successful control.
\par
\endgroup

\FloatBarrier

\section{Related Work}
\label{sec:related}

\paragraph{World models and executable representations.}
World models support decision making by predicting the consequences of actions. Dyna combines
real interaction with model-generated experience for planning and learning~\citep{sutton1991dyna}.
Neural world models support imagined rollouts, model-predictive control, and tree
search~\citep{ha2018worldmodels,chua2018,schrittwieser2020muzero}, while Dynalang incorporates
language as a predictive modality in a multimodal latent model~\citep{lin2024dynalang}. These
approaches encode learned dynamics in neural parameters and latent representations. Programs
provide an alternative representation that exposes transitions as functions a planner can execute
and test directly.

Language-guided program induction provides a foundation for constructing such models. Parsel
combines natural-language decompositions with execution-based testing~\citep{zelikman2023parsel},
and Hypothesis Search translates language hypotheses into programs verified on input-output
examples~\citep{wang2024hypothesis}. For interactive environments, WorldCoder learns Python dynamics
models and repairs them against observed counterexamples~\citep{tang2024worldcoder}; GIF-MCTS
searches over candidate models through generation, improvement, and repair guided by descriptions,
tests, and trajectories~\citep{dainese2024gifmcts}. Code World Models for General Game Playing
constructs transition, legality, and termination functions from rules and trajectories to support
Monte Carlo tree search~\citep{lehrach2025codeworld}. Code-as-World extends executable modelling to
physical reasoning through a loop of program proposal, execution, rendering, and
verification~\citep{wang2026codeasworld}. Together, these works establish language-guided construction
and execution-based refinement as useful mechanisms for learning explicit models.

Several ARC-AGI-3 systems closely connect model construction, verification, and action selection.
Rodionov's baseline1 maintains, simplifies, and verifies a persistent Python model before planning
through it~\citep{rodionov2026executable}. Its subsequent component study finds exact verification
consistently useful within the tested configurations, but does not find that requiring an executable
model uniformly improves action performance~\citep{rodionov2026components}. OPINE-World maintains
natural-language hypotheses, synthesises an object-centric executable, and uses ontology error to
direct exploration; candidate programs must reproduce recorded transitions
exactly~\citep{opineworld}. Twin similarly gates model-based action on replay of the interaction
history~\citep{skoutnev2026twin}. Tycho examines when an agent should construct, repair, use, or
bypass an executable model, showing that better transition replay need not yield better decisions
when objectives are misidentified or models are used inappropriately~\citep{lehmann2026tycho}.
These results motivate examining how a learned model is maintained and used, alongside its
predictive accuracy.

\paragraph{Language rules and reflective memory.}
Textual memory offers another way for frozen LLM agents to learn from interaction. Reflexion stores
linguistic reflections on task feedback to improve later decisions~\citep{shinn2023reflexion}.
AutoManual induces and organises environment rules into reusable
manuals~\citep{chen2024automanual}. WALL-E updates and prunes symbolic rules by comparing predicted
and observed trajectories, using those rules to align an LLM world model with its
environment~\citep{zhou2024walle}. These approaches range from retaining advice about past decisions
to maintaining explicit descriptions of environment behaviour, establishing language as an editable
medium for both experience and rule knowledge.

The Memento series develops reflective memory from episodic adaptation~\citep{zhou2025memento}
to stateful reflective learning, which links memory writing and reading to policy evaluation and
improvement~\citep{wang2025memento2}, and then to reusable executable
skills~\citep{zhou2026mementoskills}. \codemodel extends this progression to model-side semantic
memory: a persistent rulebook and its executable realisation jointly represent a world-model
hypothesis. Reflection can revise an environment rule or repair its implementation while retaining
the rule. The LLM assesses code--rulebook fidelity, and replay checks the executable against recorded
transitions. Language rules, executable artefacts, reflection, and replay are established components;
we organise them around a persistent environment specification supporting both semantic revision
and implementation repair.

\paragraph{Model uncertainty and exploration.}
Version spaces formalise the ambiguity left by finite evidence~\citep{mitchell1977version,mitchell1982},
and minimum description length provides a simplicity preference among consistent
hypotheses~\citep{rissanen1978modeling}. Model-based reinforcement learning represents uncertainty
through probabilistic dynamics models or ensembles~\citep{kurutach2018,chua2018}. Posterior sampling
and related methods retain a sampled hypothesis or value function across an episode to support
temporally coherent exploration~\citep{osband2013,osband2016,luvanroy2017}. Our population extension
applies this principle by maintaining multiple world models in parallel. Each member pairs a
rulebook with an executable, and all members are checked against shared interaction evidence.
Sampling a member for planning allows different models to guide exploration. The population is a
computational approximation to the remaining version space, rather than a calibrated Bayesian
posterior; its purpose is to reduce dependence on a single early model construction.

\paragraph{Continual agent adaptation.}
Broader approaches to agent adaptation treat external software and workspaces as editable learning
state. Workspace Optimization instantiates this view in DreamTeam, whose specialised agents build
models, probe hypotheses, and plan~\citep{sarafian2026workspace}. Continual Harness supports
reset-free revision of prompts, subagents, skills, and memory~\citep{continualharness}. NOOA combines
a Python object-oriented agent interface with persistent executable modules, prediction-error-driven
refinement, and cross-level memory in ARC-AGI-3~\citep{nvidialabs2026nooa}. Belief-Calibrated
Optimization maintains a language world model of how evaluation responds to changes in an agent's
prompt and scaffolding~\citep{chen2026bco}. Its model guides edits to the agent, whereas \codemodel
predicts transitions within the environment to support planning and control. These models support
complementary forms of agent improvement.

ARC-AGI-3 provides a shared setting for studying these approaches under unfamiliar rules and a
human-derived action budget~\citep{arcagi3}. It also supports alternatives such as graph-based
exploration, which records observed transitions and prioritises untested actions without a generative
world model~\citep{rudakov2025graph}. Tycho and Rodionov's component study report saturation or
near-saturation of the public set with stronger frontier
models~\citep{lehmann2026tycho,rodionov2026components}. Public-set completion alone therefore cannot
isolate the contribution of a world-model architecture. Comparisons must also account for model
choice and reasoning budget, and examine action efficiency, repeated-run stability, and held-out
evaluation.

\section{Conclusion}
\label{sec:conclusion}

We introduced \textsc{Memento~3}, which extends reflective learning to explicit world models
through \codemodel. The agent records environment hypotheses in a natural-language rulebook,
compiles them into code for prediction and planning, and revises both through interaction while
keeping the underlying LLM fixed. The population extension maintains multiple world models in
parallel, each updated and verified against a shared interaction history. On ARC-AGI-3, the
single-model agent clears every level of all $\numgames$ public games and reaches the RHAE ceiling,
using \ouractions{} actions, or \actionpercentage{} of the human action count. In Atari Pong, the
learned model supports a feedback controller that wins $21\!:\!0$ in three evaluated episodes with
different openings, without LLM calls during execution. By carrying verified model updates into
subsequent planning and exploration, the agent uses what it has learned to shape the evidence for
further revisions. This feedback is the model-based route to recursive self-improvement
investigated in this work.

{\small
\bibliographystyle{plainnat}
\bibliography{CaM_refs}}

\end{document}